# Data-Efficient Semantic Segmentation of 3D Point Clouds via Open-Vocabulary Image Segmentation-Based Pseudo-Labeling

**Takahiko Furuya*,†**

* University of Yamanashi, 4-3-11 Takeda, Kofu-shi, Yamanashi-ken, 400-8511, Japan
E-mail: takahikof@yamanashi.ac.jp
† Corresponding author

**Semantic segmentation of 3D point cloud scenes is a crucial task for various applications. In real-world scenarios, training segmentation models often faces three concurrent forms of data insufficiency: scarcity of training scenes, scarcity of point-level annotations, and absence of 2D image sequences from which point clouds were reconstructed. Existing data-efficient algorithms typically address only one or two of these challenges, leaving the joint treatment of all three unexplored. This paper proposes a data-efficient training framework specifically designed to address the three forms of data insufficiency. Our proposed algorithm, called *Point pseudo-Labeling via Open-Vocabulary Image Segmentation* (*PLOVIS*), leverages an open-vocabulary image segmentation model as a pseudo label generator to compensate for the lack of training data. PLOVIS creates 2D images for pseudo-labeling directly from training 3D point clouds, eliminating the need for 2D image sequences. To mitigate the inherent noise and class imbalance in pseudo labels, we introduce a two-stage filtering of pseudo labels combined with a class-balanced memory bank for effective training. The two-stage filtering mechanism first removes low-confidence pseudo labels, then discards likely incorrect pseudo labels, thereby enhancing the quality of pseudo labels. Experiments on four benchmark datasets, i.e., ScanNet, S3DIS, Toronto3D, and Semantic3D, under realistic data-scarce conditions (a few tens of training 3D scenes, each annotated with only <100 3D points) demonstrate that PLOVIS consistently outperforms existing methods including standard fine-tuning strategies and state-of-the-art weakly supervised learning algorithms. Code will be made publicly available.**

**Keywords**: 3D point cloud, semantic segmentation, deep learning, pseudo labeling

## 1. Introduction

Semantic segmentation of three-dimensional (3D) point cloud scenes is a fundamental task in computer vision and supports a wide range of applications, including autonomous driving, robotics, augmented reality, and facility maintenance. In recent years, deep learning-based methods for 3D point cloud segmentation have been studied extensively [1], [2]. In particular, advances in deep neural network (DNN) architectures [3], self-supervised pretraining algorithms [4], and data augmentation techniques [5] have steadily improved segmentation accuracy.

Most existing methods for 3D point cloud segmentation assume idealized training data conditions. That is, DNNs are trained on highly curated benchmark datasets (e.g., [6], [7], [8]) that contain a large number of 3D point cloud scenes, each densely annotated by humans. In addition, some methods (e.g., [9], [10], [11]) leverage the original 2D image sequences used to reconstruct 3D point clouds to generate additional supervision using vision foundation models. However, in real-world application scenarios, training a segmentation model often confronts three forms of data insufficiency:

- **Scarcity of scenes.** The number of 3D point cloud scenes available for training is often limited because scene acquisition requires substantial time and human effort. Especially in constrained environments such as factory plants and construction sites, the number of scenes that can be captured is inherently restricted, making it impractical to prepare hundreds or thousands of point cloud scenes for training.
- **Scarcity of annotations.** A limited number of semantic labels are typically assigned to each 3D point cloud scene. Since each scene may contain an enormous number (e.g., 1M) of 3D points, manually annotating every 3D point requires prohibitive human labor. As a result, constructing

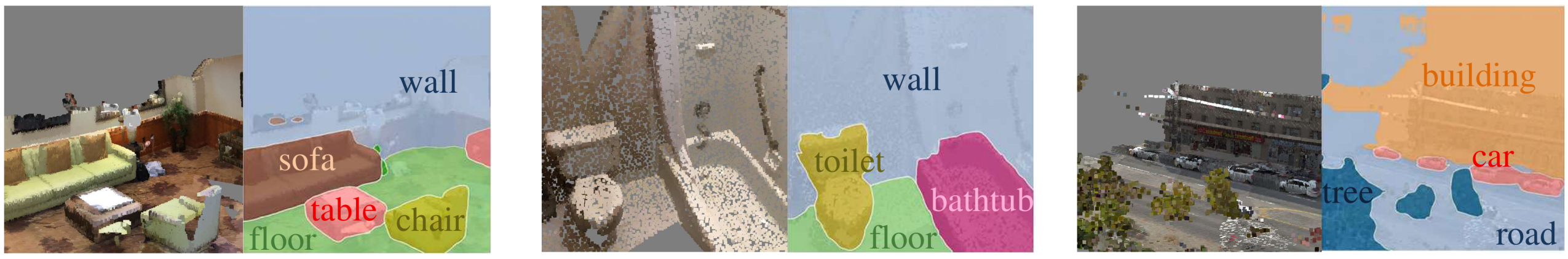

**Fig. 1.** Segmentation results produced by the OVIS model (DeCLIP [16]) for three rendered point cloud images from the ScanNet [6] and Toronto3D [17] datasets. Despite the unnatural appearance of the images, the OVIS model achieves reasonably accurate semantic segmentation, suggesting potential usability of the OVIS model as a pseudo-label generator.

densely annotated training datasets is usually unrealistic.

- **Absence of image sequences.** One of the common approaches to creating 3D point clouds is to reconstruct them from 2D natural image sequences using Structure-from-Motion (SfM) with Multi-View Stereo (MVS), or Visual SLAM. Jointly exploiting 3D point clouds and their corresponding 2D images can facilitate the training of highly accurate segmentation models. However, in real-world problems, 2D image sequences are often discarded after 3D point clouds have been generated to reduce storage costs. Consequently, multi-modal training approaches that rely on 2D-3D correspondence are not always feasible.

Aiming at data-efficient training of 3D point cloud segmentation models, various approaches have been developed [12], including weakly supervised learning, domain transfer learning, and multimodal learning. However, most existing studies address only one or two of the aforementioned three data insufficiencies, leaving the simultaneous treatment of all three largely underexplored. For example, weakly supervised learning [13] (also referred to as semi-supervised learning) primarily targets the scarcity of annotations. Weakly supervised learning employs sparsely annotated point cloud scenes in which only a subset of 3D points is labeled by humans. While this framework achieves low annotation costs, it presupposes the availability of a large number of point cloud scenes for training. Domain transfer learning ([14], [15]) addresses the scarcity of scenes and labels in the target domain by leveraging knowledge from a source domain. However, training on the source domain relies on large-scale, densely annotated point cloud scene datasets. Multimodal learning ([9], [10], [11]) achieves accurate point cloud segmentation under the condition of scarce scenes and annotations by utilizing rich knowledge from external domains such as images and text. However, these existing methods require access to large quantities of natural image sequences that are aligned with the point cloud scenes for 2D-3D joint training.

Our goal in this paper is to achieve data-efficient and accurate semantic segmentation of 3D point clouds. In particular, we aim to address three data limitations simultaneously: scarcity of scenes, scarcity of annotations, and absence of image sequences. To this end, we fully leverage an Open-Vocabulary 2D Image Segmentation (OVIS) model in combination with a Point Cloud Foundation (PCF) model. **Fig. 1** motivates the use of an OVIS model. **Fig. 1** presents the segmentation results obtained by applying the OVIS model, DeCLIP [16], to 2D images rendered from 3D point clouds. Despite the partially corrupted and pixelated renderings, the OVIS model produces reasonable segmentation results, suggesting its strong generalization capabilities. This observation motivates us to explore whether OVIS-based pseudo-labeling can alleviate the challenges of scarce annotations and unavailable image sequences. Also, to address the remaining challenge of scene scarcity, we leverage a PCF model that has learned segmentation-ready, discriminative features through large-scale self-supervised pretraining.

Our proposed algorithm, called *Point pseudo-Labeling via Open-Vocabulary Image Segmentation* (*PLOVIS*), employs the OVIS model as a pseudo labeler to 3D points. Pseudo-labeling is accomplished by rendering a 3D point cloud into a 2D image followed by pixel-level zero-shot classification using the OVIS model. Each pseudo label is paired with its corresponding pointwise feature extracted by the PCF model [4], and the resulting pairs serve as augmented training data.

However, naively using all pseudo labels generated via OVIS for training is not effective for two reasons. First, the pseudo labels can be noisy due to prediction uncertainty and errors. Noisy pseudo labels, having low confidence and/or incorrect class assignment, impede training. Second, the class distribution among pseudo labels can be imbalanced. In general, the number of 3D points belonging to each semantic class varies substantially due to differences in object size within point cloud scenes. For example, in an indoor scene, background objects such as floors and walls contain far more 3D points than foreground objects such as chairs. The pseudo labels generated by OVIS inherit this class imbalance, biasing training toward majority classes and overlooking minority classes.

To effectively learn from noisy and imbalanced pseudo labels, PLOVIS adopts two strategies: (1) a two-stage pseudo label filtering mechanism guided by confidence and loss, and (2) training with a class-balanced memory bank. The filtering process removes pseudo labels that are deemed noisy due to low confidence or possibly incorrect class assignment, retaining only reliable pseudo labels. Meanwhile, the class-balanced memory bank facilitates balanced learning across all semantic classes.

The proposed algorithm offers three strengths. First, PLOVIS is data-efficient. Training dataset of PLOVIS consists only of a small number of sparsely annotated point cloud scenes (e.g., a few dozen scenes, each annotated with only <100 points). 2D images required as input to the OVIS model are rendered directly from the point clouds, eliminating the need for the image sequences. Second, PLOVIS is robust to overfitting since it optimizes only a lightweight segmentation head, while both the PCF and OVIS models remain frozen during training. Such a framework is particularly well-suited to the data-scarce settings in this study. Third, PLOVIS is flexible since it is built upon PCF and OVIS models trained separately. This modular design allows various combinations of models. While this paper uses Sonata [4] as the PCF model and DeCLIP [16] as the OVIS model, these components can be replaced with stronger models that will be developed in the future.

Our work is closely related to weakly supervised learning [13], as applying such methods to a limited number of training scenes aligns with the three data limitations considered in this paper. Therefore, we employ existing weakly supervised point cloud segmentation algorithms as competitors. Our experiments using both indoor and outdoor point clouds demonstrate that PLOVIS achieves higher segmentation accuracy than existing state-of-the-art methods as well as standard fine-tuning strategies.

Contribution of this paper can be summarized as follows:

- We tackle 3D point cloud semantic segmentation under three realistic constraints on training data: scarcity of scenes, scarcity of annotations, and absence of image sequences.
- We propose PLOVIS, a data-efficient training framework that leverages OVIS-based pseudo-labels enhanced by a filtering mechanism and a balanced memory bank.
- We demonstrate that PLOVIS achieves higher segmentation accuracy than state-of-the-art weakly supervised learning methods on both indoor and outdoor point cloud segmentation under data-scarce conditions.

The rest of this paper is organized as follows. Section 2 reviews related work. Section 3 presents the proposed algorithm in detail. Section 4 describes the experimental setup and presents results and discussions. Finally, we conclude the paper with a summary and directions for future work.

## 2. Related work

### 2.1. Data-efficient semantic segmentation of 3D point clouds

#### 2.1.1. Motivation for data-efficient training

Research on semantic segmentation of 3D point clouds has advanced significantly ([1], [2]). In recent years, end-to-end DNN architectures that directly process 3D point cloud data have become the standard approach for semantic segmentation of 3D point clouds (e.g., [3], [18], [19], [20]). In particular, Transformer-based architectures such as Point Transformer V3 [3] have achieved remarkable segmentation accuracy on large-scale point cloud datasets. The self-attention mechanism within Transformer effectively captures the complex 3D structure of scenes by jointly modeling local neighborhood relationships and global contexts. These DNNs are typically trained in a fully supervised manner by using large-scale, densely annotated training datasets such as ScanNet [6], S3DIS [7], nuScenes [8], and SemanticKITTI [21].

However, as discussed in Section 1, such ideal training datasets are not always available in real-world application scenarios due to enormous cost and human effort for dataset construction. Training the aforementioned DNNs on a limited number of sparsely annotated point clouds may lead to severe overfitting. Consequently, there is a need for data-efficient learning approaches that can effectively train segmentation models with limited amount of training data.

#### 2.1.2. Approaches to data-efficient 3D point cloud segmentation

Various approaches have been proposed for effective training of segmentation models under low-data regimes [12]. These approaches can be roughly categorized into weakly supervised learning, domain transfer learning, and multi-modal learning.

**Weakly supervised learning.** Unlike fully supervised learning, weakly supervised learning [13], or semi-supervised learning, assumes that only a small subset of 3D points is labeled, while the vast majority remain unlabeled. Recent approaches employ pseudo-labeling ([22], [23], [24]) or improved loss functions ([25], [26]). Pseudo labels are typically generated by training a DNN on a small number of labeled points and then using the DNN to predict labels for the unlabeled points. The large set of pseudo labels is used together with the small set of human-annotated true labels for further training. ERDA by Tang et al. [22] employs this pseudo-labeling strategy. To bridge the gap between the noisy pseudo labels and the predictions by the segmentation DNN, ERDA performs label distribution alignment based on entropy and KL divergence. PointMatch [23] applies consistency regularization [27] to enforce consistent predictions across different data augmentations, effectively propagating sparse true labels to neighboring points and improving the quality of pseudo labels. RAC-Net [24] also adopts the consistency regularization to estimate the reliability of pseudo labels and weight them according to confidence, thereby reducing the negative impact of noisy pseudo labels. Alternatively, rather than relying on pseudo labels, some approaches improve the loss function to enable effective learning from a small number of true labels. DG-Net by Gao et al. [25] employs a mixture of von Mises-Fisher distributions as a prior and introduces a loss that guides feature distribution alignment toward this prior, stabilizing learning from limited number of labels. AAD-Net by Pan et al. [26] addresses learning from sparse and inhomogeneously distributed annotations by adopting a multiplicative dynamic entropy loss to correct gradient bias, thereby achieving robust training.

These methods have shown that effective learning is possible even from highly sparse annotations (e.g., 0.01% of all 3D points). However, they still require a large number of point cloud scenes for training, leaving the challenge of achieving high accuracy from limited number of training scenes. In contrast, our work differs from the previous studies in that ours uses only a small number of sparsely annotated point cloud scenes for training.

**Domain transfer learning.** Domain transfer learning, or domain adaptation, is a technique that adapts knowledge learned from a label-rich source domain to a label-scarce target domain. Prior work on 3D point cloud segmentation ([14], [15]) covers various transfer scenarios, including synthetic-to-real, clear-to-adverse weather, and city-to-city. Representative methods employ adversarial learning [28] or pseudo-labeling [29] to bridge the discrepancies of feature distribution between domains. While domain transfer learning is powerful in that it can leverage abundant source-domain knowledge, it requires large amounts of labeled point clouds as source data, making it fundamentally different from the data efficiency targeted in our study.

**Multimodal learning.** Multimodal learning, or cross-modal learning, exploits multiple data modalities such as 2D images or texts. Some methods utilize 2D image modality for 3D point cloud segmentation [12]. Rich visual information contained in natural 2D images is useful to detect objects that are difficult to identify from 3D point clouds alone. Yan et al. [9] and Zeid et al. [11] introduce cross-modal knowledge distillation. They use powerful 2D image features extracted by pretrained vision foundation models such as DINOv2 [30] to train a point cloud encoder for generalizable 3D shape features. Kweon et al. [31], Deng et al. [32], and Dong et al. [10] employ Segment Anything Model (SAM) [33] and CLIP [34] to assign pseudo labels to image pixels associated with 3D points, thereby generating pointwise pseudo labels and enabling label-efficient point cloud encoder training. Kweon et al. [35] and Duan et al. [36] propose joint learning frameworks that employ both distillation and pseudo-labeling.

The abovementioned multimodal approaches have successfully trained point cloud encoders with high generalization capability by leveraging image/text domain knowledge. However, their training relies heavily on 2D image sequences that are precisely registered with 3D point clouds. Therefore, these methods cannot be applied in scenarios where image sequences are unavailable. Although our work also utilizes a vision-language model (i.e., an OVIS model), it differs from existing multimodal approaches in that we assume no access to original image sequences and instead use images rendered directly from 3D point clouds.

## 2.2. Foundation models for 3D point clouds

A foundation model is a large DNN pretrained on massive and diverse datasets, enabling effective adaptation to various downstream tasks through fine-tuning, even with limited training samples. Motivated by the recent success of powerful foundation models for 2D image [30] and text [37], large-scale self-supervised pretraining of Point Cloud Foundation (PCF) models has become active research area [38]. Point-BERT [39] and its variants (e.g., [40], [41]) pretrain point cloud Transformer via a pretext task called masked point modeling, which requires the DNN to reconstruct masked local regions of 3D objects. Wu et al. [42] extended masked point modeling from the object level to the scene level to develop DNNs for point cloud scene analysis. Subsequently, Wu et al. [4] developed a powerful PCF model called Sonata by pretraining a large-scale point cloud Transformer [3] via a self-distillation framework [30]. Concerto [43] is a 2D-3D joint pretraining framework built upon Sonata, combining intra-modal self-distillation with cross-modal knowledge distillation from a vision foundation model [30]. DoReMi [44] is a PCF model that employs a mixture-of-experts architecture to effectively learn geometric features that are both shared and distinct across multiple datasets. Recently proposed PointINS [45] employs instance-level self-distillation to learn instance-aware geometric features.

These studies on PCF models have demonstrated excellent accuracy on the downstream task of 3D point cloud segmentation. Note that, however, these PCF models primarily assume that the pretrained model will be fine-tuned by using a large amount of densely annotated point clouds. Our work leverages the expressive pointwise features extracted by a frozen PCF model and associates these features with pseudo labels predicted by an OVIS model, aiming at improved data efficiency in segmentation training.

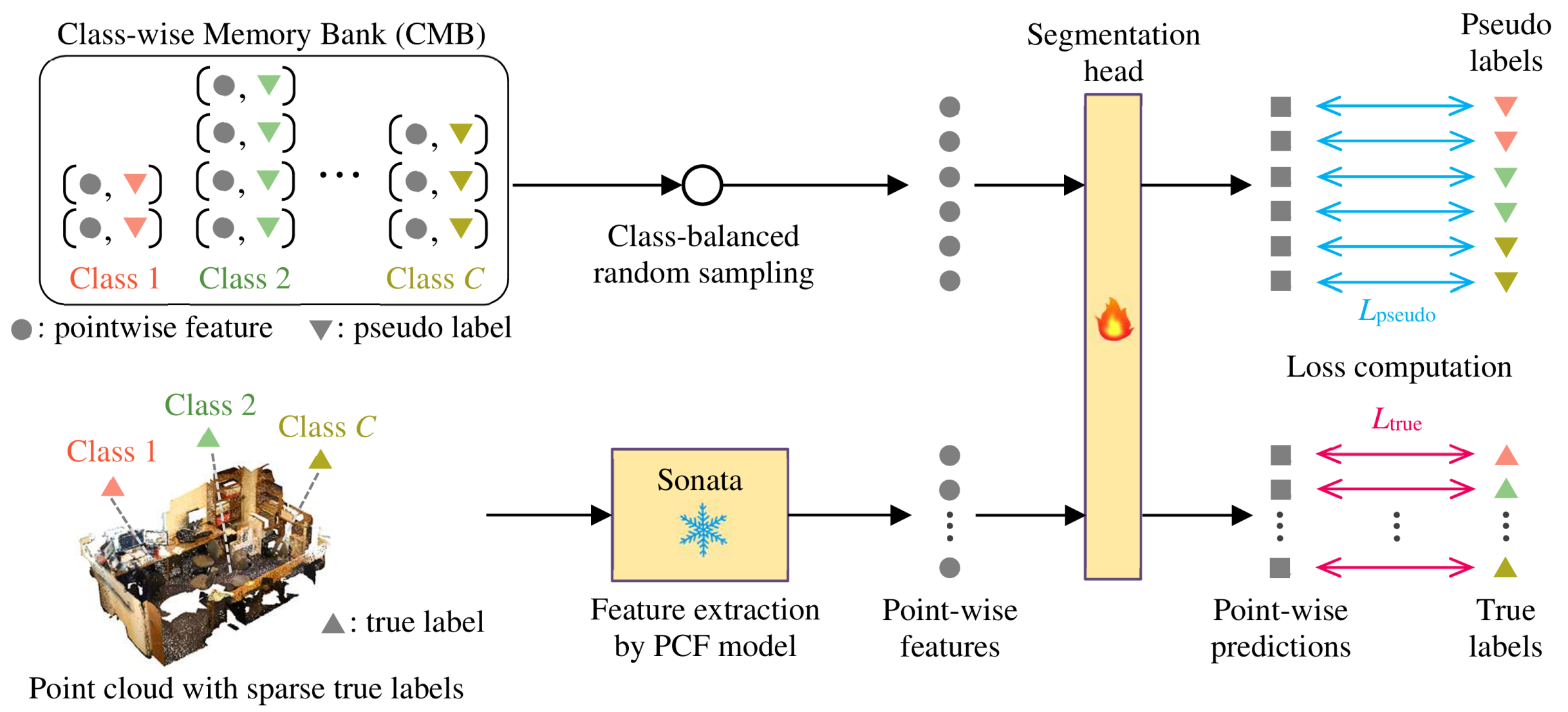

**Fig. 2.** Training framework of the proposed PLOVIS algorithm. The lightweight segmentation head is jointly trained by using two types of supervision: a small number of human-annotated true labels and a large number of pseudo labels predicted by the OVIS model. The use of a memory bank promotes class-balanced learning.

### 2.3. Open-vocabulary 2D image segmentation

Recent advances in large-scale vision-language foundation models have rapidly driven the development of Open-Vocabulary Image Segmentation (OVIS), which predicts pixel-level labels for arbitrary classes described by texts ([46], [47]). In particular, the advent of CLIP [34], a model pretrained on large-scale image-text pairs, dramatically enhanced the generalization capability of OVIS. LSeg [48] realized zero-shot, open-vocabulary 2D image segmentation by aligning pixel embeddings from an image encoder with class-name embeddings from CLIP's text encoder. OpenSeg [49] and OVSeg [50] proposed frameworks that align patch-level embeddings, which is extracted from local regions of an image, with text embeddings instead of pixel-level embeddings. CAT-Seg [51] treats the cosine similarity between CLIP image feature maps and text embeddings as a cost volume and aggregates the volume both spatially and semantically to improve OVIS accuracy for classes unseen during training. DeCLIP by Wang et al. [16], which we use in this study, decouples an output from CLIP's final self-attention module into content and context features for distillation, optimizing vision-language alignment while enhancing spatial consistency. DeCLIP demonstrated that integrating this decoupled distillation strategy into CAT-Seg boosts accuracy of open-vocabulary semantic segmentation. Other efforts to improve OVIS models include methods (e.g., [52]) that leverage class-agnostic segmentation foundation models such as SAM [33]. More recently, SAM 3 [53], which integrates a CLIP-like vision-language model called Perception Encoder with SAM and enables text-prompted segmentation, has also been proposed.

Note that the objective of our work is not open-vocabulary 3D point cloud segmentation ([54], [55], [56]). Instead, we utilize the OVIS model as a pseudo label generator to compensate for the lack of training data.

## 3. Proposed algorithm

### 3.1. Overview of PLOVIS

**Figure 2** illustrates the training framework of PLOVIS. PLOVIS optimizes only a lightweight segmentation head implemented as a multi-layer perceptron (MLP). The segmentation head parameters are updated using two types of supervision signals, i.e., human-annotated ground-truth labels (true labels) and OVIS-based pseudo labels. The lower part of **Fig. 2** depicts the training pipeline with true labels. An input 3D point cloud that is sparsely annotated by humans are transformed into a set of pointwise features by using a pretrained, frozen PCF model. The pointwise features are then processed by the segmentation head to obtain predictions for each point. The upper part of **Fig. 2** illustrates the training pipeline with pseudo labels. We construct a storage buffer called a Class-wise Memory Bank (CMB) to store diverse pairs of pointwise features and pseudo labels. By sampling training pairs

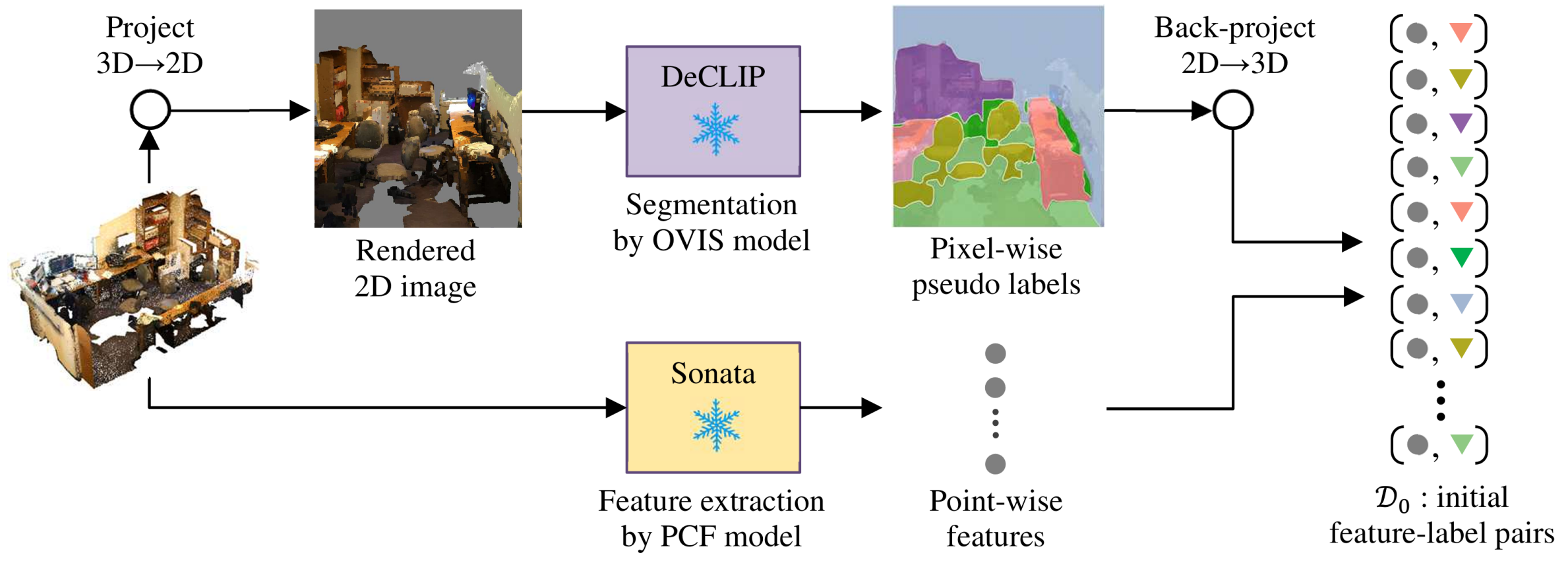

**(a) Pseudo label generation.**

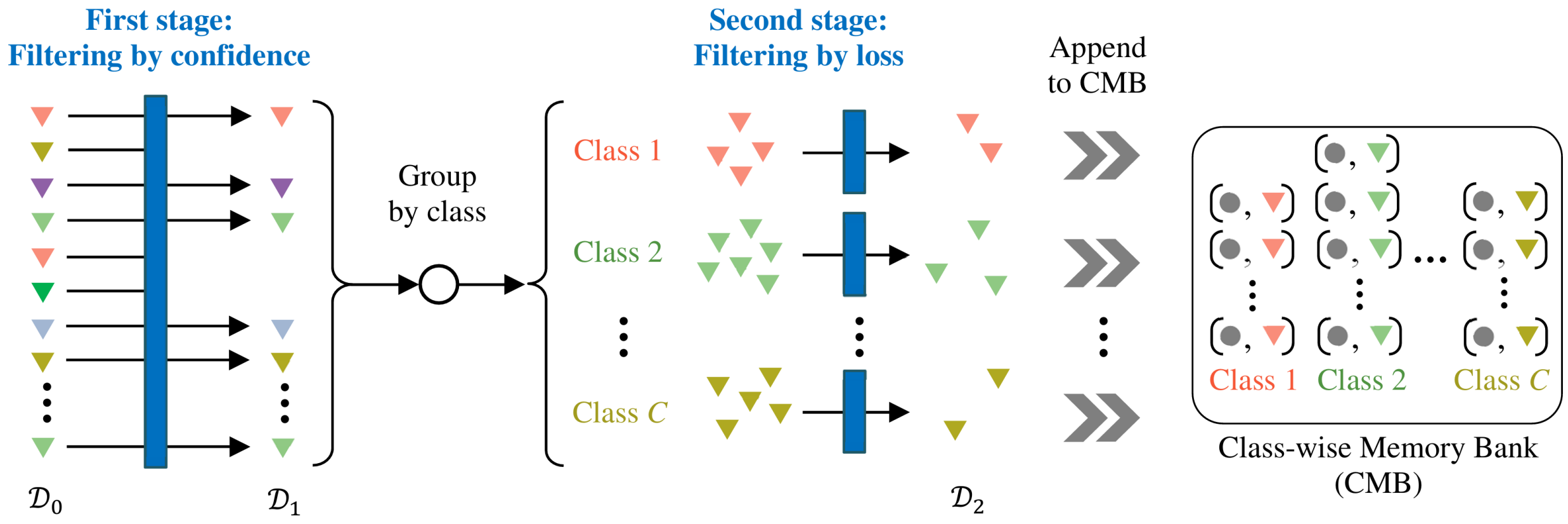

**(b) Two-stage pseudo label filtering and updating CMB.**

**Fig. 3.** Procedures of pseudo label generation, pseudo label filtering, and CMB update. (a) The 3D points in the training point cloud are assigned pseudo labels by the OVIS model in the 2D domain and then paired with pointwise features extracted by the PCF model. (b) The pseudo labels are refined through a two-stage filtering process that removes unconfident and potentially incorrect pseudo labels. The refined pseudo labels are then appended to the CMB and used for training.

from CMB in a class-balanced manner, we compensate for the scarcity of true labels while mitigating the training bias toward majority classes.

**Figure 3** illustrates the processes for pseudo label generation, pseudo label filtering, and subsequent CMB update. As shown in **Fig. 3a**, a training 3D point cloud is first rendered into a 2D image and then pseudo-labeled by the OVIS model. The pixel-wise pseudo labels are back-projected from 2D to 3D and associated with the corresponding 3D point features, producing initial feature-label pairs. While one could directly use the initial pairs for training, these pseudo labels are noisy due to uncertain and/or incorrect predictions. Noisy pseudo labels may impede the training of the segmentation head. We thus propose a two-stage filtering mechanism, shown in **Fig. 3b**, to selectively retain reliable pseudo labels. The first filtering stage aims to filter out uncertain pseudo labels generated with low confidence. For example, pseudo labels assigned to 3D points on indiscernible objects or object boundaries are filtered out in this stage. The second filtering stage aims to remove pseudo labels that are confident yet likely incorrect. Inspired by prior work on learning with noisy labels ([57], [58], [59]), we use the loss values computed by the segmentation head during training as an indicator for filtering.

Note that the abovementioned processes are contained within the PLOVIS training loop. That is, each training step involves pseudo label generation (**Fig. 3a**), pseudo label filtering followed by CMB update (**Fig. 3b**), and training of the segmentation head (**Fig. 2**).

### 3.2. Pointwise prediction using PCF model

We first describe the inference path depicted in the lower part of **Fig.2**. Let $X = \{\mathbf{x}_i \mid i = 1, \dots, N\}$ denote a 3D point cloud scene contained in a training mini-batch, where $N$ is the number of 3D points within the scene and each $\mathbf{x}_i$ represents the coordinates and attributes (e.g., color and normal) of the $i$-th point. We assume that ground-truth labels, each denoted by $\mathbf{y}_i$, are available for only a very small subset of the 3D points in $X$. Each $\mathbf{y}_i$ is represented as a $C$-dimensional one-hot vector, where $C$ is the number of candidate classes for semantic segmentation.

We feed $X$ into the PCF model to obtain the set of pointwise features $f(X) = \{\mathbf{f}_i \mid i = 1, \dots, N\}$. This paper uses Sonata [4] as the self-supervisedly pretrained encoder $f$ with its parameters frozen during training. In the case of Sonata, each pointwise feature vector $\mathbf{f}_i$ has 1,232 dimensions. The pointwise features are independently processed by the segmentation head $h$ to produce $C$-dimensional pointwise logits: $\mathbf{h}_i = h(\mathbf{f}_i)$. To mitigate overfitting, we implement the function $h: \mathbb{R}^{1232} \rightarrow \mathbb{R}^C$ as a simple three-layer MLP with batch normalization [60] and ReLU activation. The logits are passed through the softmax function to obtain the $C$-dimensional pointwise prediction vector $\hat{\mathbf{y}}_i$:

$$\hat{\mathbf{y}}_{ic} = \frac{\exp(\mathbf{h}_{ic})}{\sum_{j=1}^{C} \exp(\mathbf{h}_{ij})} \quad (1)$$

In Eq. (1), $\hat{\mathbf{y}}_{ic}$ and $\mathbf{h}_{ic}$ denote $c$-th element of $\hat{\mathbf{y}}_i$ and $\mathbf{h}_i$, respectively.

### 3.3. Pseudo-labeling using OVIS model

To compensate for the scarcity of true labels, we perform pseudo-labeling by using the OVIS model. The processing pipeline is illustrated in **Fig. 3a**. During training, the 3D point cloud scene $X$ is converted into a 2D image $\mathbf{I}_X$ by projecting $X$ onto a 2D image plane. The camera position is randomly selected from the four upper vertices of the axis-aligned bounding box of $X$. The rendering spatial resolution is fixed at 768×768 pixels to match the input requirements of DeCLIP [16]. When multiple 3D points are projected onto the same pixel, the point closest to the camera is associated with that pixel. For each pixel, an $s$×$s$ square region centered at that pixel is filled with the color of the corresponding 3D point. The size $s$ decreases linearly from 6 pixels to 2 pixels as the distance between the point and the camera increases. The background color is set to gray. As a result, we obtain a pixelated, colored 2D image $\mathbf{I}_X \in \mathbb{R}^{768 \times 768 \times 3}$, as exemplified in **Fig. 1**.

The image $\mathbf{I}_X$ is then input to the OVIS model along with a set of candidate class text prompts, producing a 3D tensor containing pixel-wise logits: $\hat{\mathbf{I}}_X = g(\mathbf{I}_X, T)$, $\hat{\mathbf{I}}_X \in \mathbb{R}^{768 \times 768 \times C}$. We use DeCLIP as the function $g$, which is one of the state-of-the-art OVIS models. $T$ is a set of text prompts, each created by concatenating the template text "a photo of a" with a candidate class name or its synonym (e.g., bookshelf). Each logit vector $\mathbf{g}_i \in \mathbb{R}^C$ contained in $\hat{\mathbf{I}}_X$ is converted into a pseudo label vector $\tilde{\mathbf{y}}_i$ by using a temperature-scaled softmax function:

$$\tilde{\mathbf{y}}_{ic} = \frac{\exp(\mathbf{g}_{ic}/\tau)}{\sum_{j=1}^{C} \exp(\mathbf{g}_{ij}/\tau)} \quad (2)$$

In Eq. 2, $\tau$ is a hyperparameter fixed at 0.2 in this paper. The pixels are then back-projected into 3D space. The pseudo label $\tilde{\mathbf{y}}_i$ of each pixel is paired with the pointwise feature $\mathbf{f}_i$ of the corresponding 3D point to form a feature-label pair $(\mathbf{f}_i, \tilde{\mathbf{y}}_i)$. Pointwise features are extracted by using the method described in Section 3.2. Pseudo labels of pixels that do not correspond to any 3D point (i.e., background pixels) are discarded. We define the resulting initial set of training pairs as $\mathcal{D}_0 = \{(\mathbf{f}_i, \tilde{\mathbf{y}}_i) \mid i = 1, \dots, M\}$ where $M$ is the number of 3D points labeled by the OVIS model.

### 3.4. Two-stage pseudo label filtering

We apply a two-stage filtering procedure shown in **Fig. 3b** to the initial training pairs $\mathcal{D}_0$ to remove noisy pseudo labels that are either highly uncertain or likely incorrect, thereby obtaining cleaner, more reliable feature-label pairs.

#### 3.4.1. First stage: filtering by confidence

The first stage aims at removing uncertain, low-confidence pseudo labels. As shown in **Fig. 4**, some regions of the rendered point cloud images are difficult to recognize even for humans due to incomplete object shapes, poor lighting conditions, occlusions, or ambiguous object boundaries. For pixels in such regions, the OVIS model often generates low-confidence pseudo labels characterized by flat class probability distributions. We thus use the entropy of the pseudo label vector as a numerical indicator for confidence-based filtering:

$$H(\tilde{\mathbf{y}}_i) = -\sum_{j=1}^{C} \tilde{\mathbf{y}}_{ij} \log(\tilde{\mathbf{y}}_{ij}) \quad (3)$$

A larger value of $H(\tilde{\mathbf{y}}_i)$ indicates a flatter class probability distribution, while a smaller value indicates a sharper distribution close to a one-hot vector. Assuming that $\mathcal{D}_0$ contains both high-confidence and low-confidence pseudo labels, we retain the training pairs with small entropy values to obtain the first-stage filtering result: $\mathcal{D}_1 = \{(\mathbf{f}_i, \tilde{\mathbf{y}}_i) \mid i \in I_r\}$, where $I_r$ is the set of indices whose $H(\tilde{\mathbf{y}}_i)$ values fall within the bottom $r$ % percentile, meaning that $(100 - r)$ % of pairs in $\mathcal{D}_0$ are removed. The retention rate $r$ is a hyperparameter (e.g., $r = 50$).

**Figure 4** exemplifies the filtering results. The red points in **Fig. 4** represent 3D points removed by the first filtering stage, i.e., $\mathcal{D}_0 \setminus \mathcal{D}_1$. We can observe that 3D points associated with low-entropy pseudo labels tend

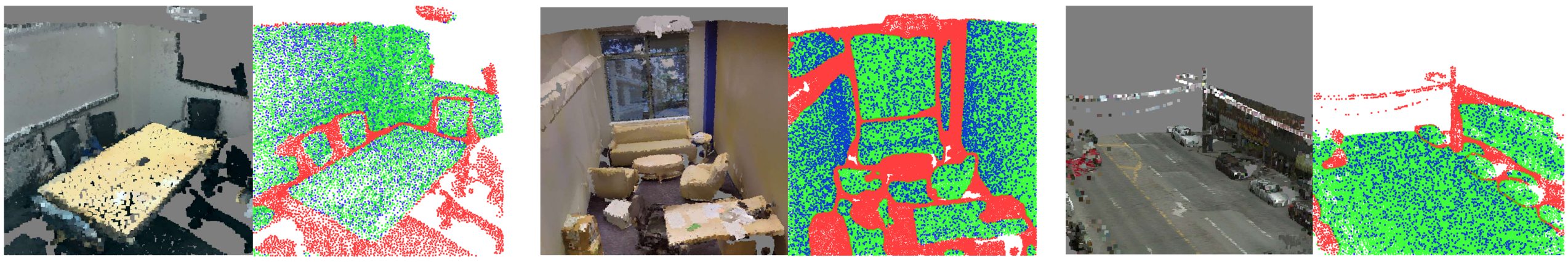

**Fig. 4.** Examples of two-stage pseudo label filtering results. Red points denote 3D points removed in the first stage (filtering by confidence), which tend to lie near object boundaries or on hard-to-recognize objects. Blue points denote 3D points filtered out in the second stage (filtering by loss), which are estimated to be noisy for training. Green points are the 3D points associated with clean pseudo labels retained after filtering, which are accumulated in the CMB and used to train the segmentation head.

to appear in hard-to-recognize regions and on object boundaries.

#### 3.4.2. Second stage: filtering by loss

Even after removing low-confidence pseudo labels in the first stage, $\mathcal{D}_1$ may still contain pseudo labels that are confident yet incorrect. The second stage targets the removal of such erroneously pseudo-labeled 3D points caused by mislabeling by OVIS. Several prior studies on learning with noisy labels ([57], [58], [59]) have observed that the training loss for incorrectly labeled samples remains high even in the later epochs of training. This observation suggests that the loss values computed by the DNN during training can serve as a numerical indicator for estimating the degree of label error. We incorporate this loss-driven idea into the second stage of our pseudo label filtering.

Specifically, we first partition $\mathcal{D}_1$ into class-wise disjoint subsets according to the maximum elements within the pseudo-label vectors:

$$\mathcal{D}_1 = \bigcup_{j=1}^{C} \mathcal{D}_1^j, \quad where \quad \mathcal{D}_1^j = \left\{ (\mathbf{f}_i, \tilde{\mathbf{y}}_i) \;\middle|\; \underset{1 \le k \le C}{\operatorname{argmax}}\, \tilde{\mathbf{y}}_{ik} = j \right\} \tag{4}$$

We apply the second-stage filtering to each subset independently, ensuring that feature-label pairs are sampled from all the $C$ classes. For each feature-label pair $(\mathbf{f}_i, \tilde{\mathbf{y}}_i)$ in $\mathcal{D}_1^j$, the pointwise feature $\mathbf{f}_i$ is passed through the segmentation head and its subsequent softmax function to produce the pointwise prediction $\hat{\mathbf{y}}_i$. We then compute the training loss $l_i$, i.e., cross-entropy between $\hat{\mathbf{y}}_i$ and $\tilde{\mathbf{y}}_i$. Next, a two-component 1D Gaussian Mixture Model (GMM) clustering is applied to the set of all loss values $\left\{ l_1, \ldots, l_{|\mathcal{D}_1^j|} \right\}$ computed for the $j$-th class subset $\mathcal{D}_1^j$. The resulting two gaussians are denoted $G_{small}^j$ and $G_{large}^j$, where $G_{small}^j$ is the cluster associated with the smaller mean loss.

We collect feature-label pairs belonging to $G_{small}^j$, which are estimated to be correctly pseudo-labeled, to form the clean subset $\mathcal{D}_2^j$. We then take the union of the clean subsets across all classes to obtain the second-stage filtering result $\mathcal{D}_2$:

$$\mathcal{D}_2 = \bigcup_{j=1}^{C} \mathcal{D}_2^j, \quad where \quad \mathcal{D}_2^j = \left\{ (\mathbf{f}_i, \tilde{\mathbf{y}}_i) \;\middle|\; p\left(l_i, G_{small}^j\right) \ge 0.5 \right\} \tag{5}$$

In Eq. 5, $p\left(l_i, G_{small}^j\right)$ denotes the probability that the loss value $l_i$ belongs to the cluster $G_{small}^j$. $\mathcal{D}_2$ is added to the CMB and used for training as described in the next subsection. Since $\mathcal{D}_2$ is created at each training step, the CMB progressively accumulates diverse feature-label pairs generated from different rendering viewpoints of different point cloud scenes as training progresses. Due to GPU memory limitations, the maximum number of pairs stored in the CMB is capped at 100k per class. When this limit is exceeded, the oldest pairs are discarded to make room for new ones.

The blue points in **Fig. 4** represent 3D points removed by the second-stage filtering (i.e., $\mathcal{D}_1 \setminus \mathcal{D}_2$), while the green points are those added to the CMB (i.e., $\mathcal{D}_2$). Although the blue and green points in **Fig. 4** tend to lie on the same objects and are visually indistinguishable, our experiment will show that the pseudo labels corresponding to the green points are of higher quality than those corresponding to the blue points.

### 3.5. Training with true and pseudo labels

As illustrated in **Fig. 2**, the segmentation head $h$ is jointly trained using two types of supervisory signals: true labels annotated by humans and pseudo labels generated by the OVIS model. The loss using true labels is computed as follows. First, pointwise features $\{\mathbf{f}_i\}$ are extracted from the training point cloud scene $X$ using the method described in Section 3.2. Among these, the pointwise features that have corresponding human-annotated labels are selected to form the training pair set $\mathcal{D}_{true} = \{(\mathbf{f}_i, \mathbf{y}_i)\}$. We assume that only a few dozen (e.g., 50) 3D points are annotated per scene. After converting each pointwise feature $\mathbf{f}_i$ to a pointwise prediction $\hat{\mathbf{y}}_i$, we compute the cross-entropy loss for $\mathcal{D}_{true}$:

$$L_{true} = -\frac{1}{|\mathcal{D}_{true}|}\sum_{i=1}^{|\mathcal{D}_{true}|}\sum_{j=1}^{C}\mathbf{y}_{ij}\log(\hat{\mathbf{y}}_{ij}) \quad (6)$$

The loss using pseudo labels is computed as follows. We randomly sample up to $n$ feature-label pairs per class from the CMB to create the class-balanced training set $\mathcal{D}_{pseudo} = \{(\mathbf{f}_i, \tilde{\mathbf{y}}_i)\}$ . The CMB is initialized as empty at the beginning of training and gradually grows as training progresses. Therefore, in the early stages of training, the number of training pairs per class stored in the CMB is small and imbalanced. We thus use the first quartile of the per-class pair counts in the CMB as $n$. By setting $n$ to the count of relatively less populated classes, we approximately equalize the number of training pairs across classes. Considering GPU memory constraints, $n$ is capped at 40k once the first quartile exceeds this value. When the CMB becomes sufficiently large, $|\mathcal{D}_{pseudo}|$ becomes $nC$, which is much larger than $|\mathcal{D}_{true}|$ . After converting the pointwise features in $\mathcal{D}_{pseudo}$ to pointwise predictions, we compute their cross-entropy loss:

$$L_{pseudo} = -\frac{1}{|\mathcal{D}_{pseudo}|}\sum_{i=1}^{|\mathcal{D}_{pseudo}|}\sum_{j=1}^{C}\tilde{\mathbf{y}}_{ij}\log(\hat{\mathbf{y}}_{ij}) \quad (7)$$

The overall loss is defined as: $L = wL_{true} + (1 - w)L_{pseudo}$, where the hyperparameter $w$ balances the two loss terms. In this paper, we simply set $w$ to 0.5 to weight both terms equally during training.

We use AdamW [61] with an initial learning rate of $2\times10^{-4}$ for optimization. Each mini-batch contains six point cloud scenes. Each training point cloud is augmented by random rotation about upright axis, scaling, flipping, and point jittering, as recommended in [4]. To avoid generating visually unnatural rendered images, scene mixup is disabled. Optimization is iterated for 200 epochs.

## 4. Experiments and results

### 4.1. Experimental setup

#### 4.1.1. Datasets

We validate the effectiveness of PLOVIS under data-scarce conditions. We use four benchmark datasets for semantic segmentation of 3D point clouds: two indoor datasets, i.e., ScanNet [6] and S3DIS [7], and two outdoor datasets, i.e., Toronto3D [17] and Semantic3D [62].

ScanNet is an indoor scene dataset derived from RGB-D video sequences. Each point in a 3D point cloud scene is annotated with one of 20 semantic classes, such as wall, bed, and chair. S3DIS contains 3D point clouds of six large-scale indoor areas measured by using a handheld LiDAR sensor. Each 3D point is annotated with one of 13 semantic classes, including ceiling, door, and bookcase. Toronto3D is an urban street scene dataset acquired by a mobile laser scanning system. Each 3D point is annotated with one of 8 semantic classes, such as road, building, and car. Semantic3D is an outdoor scene dataset acquired by terrestrial laser scanners. Each point is annotated with one of 7 semantic classes, including man-made terrain, natural terrain, tree, and building. Since each point cloud in Toronto3D and Semantic3D covers a large area and contains very large number of 3D points, we crop spherical regions with a radius of 25 meters to match the point counts of ScanNet and S3DIS (approximately 1M points per scene).

The point clouds in the above datasets are densely annotated by humans. To conduct experiments under realistic conditions where both scenes and annotations are scarce, we downsample the datasets. Specifically, the maximum number of training point cloud scenes is set to 60 for ScanNet, 40 for S3DIS, 20 for Toronto3D, and 15 for Semantic3D. In addition, the maximum number of annotated points per training scene is set to 100 for all datasets. These experimental settings, i.e., scanning several tens of scenes and labeling 100 points per scene, reflect conditions that are likely to be encountered in real-world applications.

For testing data, we use the subsets provided by the original benchmarks for the indoor datasets: 312 point clouds in the validation set for ScanNet and 48 point clouds belonging to Area 5 for S3DIS. For the outdoor datasets, we use point clouds cropped from areas different from the training data: 7 scenes for Toronto3D and 15 scenes for Semantic3D.

We use mean Intersection over Union (mIoU) and mean accuracy (mAcc) as evaluation metrics. The IoU and Acc for class $c$ are defined as: $\mathrm{IoU}_c = \mathrm{TP}_c/(\mathrm{TP}_c + \mathrm{FP}_c + \mathrm{FN}_c)$ and $\mathrm{Acc}_c = \mathrm{TP}_c/(\mathrm{TP}_c + \mathrm{FN}_c)$, where $\mathrm{TP}_c$, $\mathrm{FP}_c$, and $\mathrm{FN}_c$ denote the number of true positives, false positives, and false negatives for class $c$, respectively. Intuitively, IoU quantifies the overlap between predictions and ground truth, while Acc quantifies detection coverage, or recall. The mIoU and mAcc are obtained by averaging IoU and Acc over all classes.

#### 4.1.2. Competitors

We compare the proposed PLOVIS against six existing algorithms. The first three methods adopt different fine-tuning strategies for the PCF model, i.e., MLP probing, decoder probing, and full fine-tuning. MLP probing freezes the pretrained encoder of Sonata and fine-tunes only the randomly initialized segmentation head implemented as a three-layer MLP. This setting is similar to the linear probing strategy used in [4], which applies a single-layer segmentation head. MLP probing corresponds to specifying the weight $w$ in our loss function as 1, meaning that the segmentation head is trained by using sparse human-annotated labels only. Decoder probing attaches the randomly initialized, learnable Transformer-based decoder [4] to the frozen Sonata encoder. Comparison with decoder probing validates the use of the simple, MLP-based segmentation head in PLOVIS. Full fine-tuning adjusts all parameters of both the encoder and decoder of Sonata during training. In [4], full fine-tuning achieved the highest segmentation accuracy among multiple fine-tuning strategies when a large number of densely labeled point clouds were available.

The remaining three methods adopt weakly supervised learning tailored for semantic segmentation of 3D point clouds with limited annotations. AAD-Net [26] and DG-Net [25] train segmentation DNNs using loss functions designed to avoid overfitting to a small number of ground-truth labels. ERDA [22] employs a pseudo-labeling strategy but, unlike PLOVIS, uses pseudo labels generated by the point cloud segmentation model being trained. For a fair comparison, DG-Net, AAD-Net, and ERDA use the same DNN architecture as PLOVIS: a frozen Sonata encoder followed by a learnable three-layer MLP head.

All experiments are conducted on a PC equipped with an *Intel Core i7-14700KF* CPU, 128 GB of main memory, and an *NVIDIA RTX 6000 Ada GPU*, which has 48GB of VRAM. Each experiment is run three times with different random seeds, and we report mIoU and mACC averaged over three trials. Since the standard deviation over three trials was within 1% in almost all cases, it is omitted from the report.

## 4.2. Experimental results

### 4.2.1. Comparison with existing algorithms

**Table 1** summarizes the segmentation accuracy of PLOVIS and the six competing algorithms across four datasets. In this experiment, we use 60, 40, 20, and 15 training point cloud scenes for ScanNet, S3DIS, Toronto3D, and Semantic3D, respectively, and each scene is annotated with 50 labeled points. As shown in **Table 1**, the proposed PLOVIS consistently achieves the highest mIoU and mAcc on all datasets. Specifically, PLOVIS attains mIoU scores of 60.1%, 66.0%, 68.3%, and 53.7% on ScanNet, S3DIS, Toronto3D, and Semantic3D, respectively, outperforming the best competing method by approximately 7, 2, 5, and 5 percentage points (pp). This result confirms that leveraging OVIS-based pseudo labels is an effective strategy for data-efficient 3D point cloud segmentation under the condition where training scenes and annotations are scarce.

Among the three fine-tuning strategies for the PCF model, MLP probing outperforms both decoder probing and full fine-tuning on three out of the four datasets. This result indicates that, under data-scarce conditions, training a lightweight segmentation head while freezing the powerful pretrained encoder is more effective than tuning a larger number of parameters. Full fine-tuning, which achieved the best accuracy in the Sonata paper [4] with abundant densely labeled data, suffers in our data-scarce setting, suggesting the occurrence of overfitting. The three weakly supervised learning methods, i.e., AAD-Net, DG-Net, and ERDA, yield mIoU values comparable to, or in some cases worse than, MLP probing across all four datasets. Despite their specialized loss functions or pseudo-labeling strategies, these methods provide little benefit under the data-scarce conditions assumed in this study.

In contrast, PLOVIS achieves substantially higher accuracy than all competitors. The key advantage of PLOVIS is its ability to generate a large number of diverse pseudo labels via the OVIS model and use them as supplementary supervision signals. While MLP probing and the weakly supervised methods rely solely on the limited human-annotated labels, PLOVIS compensates for the scarcity of true labels by leveraging OVIS-based pseudo labels accumulated in the CMB. As we will demonstrate in the next subsection, the two-stage filtering mechanism contributes to improving the accuracy of PLOVIS.

**Tables 2**, **Table 3**, and **Table 4** report the class-wise IoU for each dataset, providing a more detailed view of the segmentation accuracy. **Table 2** presents the per-class IoU on ScanNet. PLOVIS achieves the highest IoU for 17 out of 20 classes. Notably, for classes with relatively few training 3D points, such as bookshelf, counter, and sink, PLOVIS demonstrates substantial

**Table 1.** Comparison of segmentation accuracy across four 3D point cloud scene datasets.

| Algorithms | ScanNet | | S3DIS | | Toronto3D | | Semantic3D | |
|---|---|---|---|---|---|---|---|---|
| | mIoU [%] | mAcc [%] | mIoU [%] | mAcc [%] | mIoU [%] | mAcc [%] | mIoU [%] | mAcc [%] |
| MLP probing | 52.9 | 64.9 | 63.3 | 73.2 | 63.4 | 72.2 | 48.6 | 61.7 |
| Decoder probing | 49.9 | 62.5 | 61.6 | 72.2 | 63.3 | 74.1 | 45.6 | 59.2 |
| Full fine-tuning | 47.0 | 61.1 | 60.5 | 71.3 | 63.5 | 73.4 | 44.4 | 58.5 |
| AAD-Net | 52.8 | 65.1 | 63.4 | 72.8 | 63.1 | 71.4 | 48.1 | 61.1 |
| DG-Net | 52.9 | 67.2 | 63.4 | 72.8 | 62.7 | 70.7 | 48.4 | 63.1 |
| ERDA | 52.8 | 66.3 | 64.0 | 74.2 | 63.1 | 72.3 | 48.0 | 61.9 |
| PLOVIS (ours) | **60.1** | **73.7** | **66.0** | **77.6** | **68.3** | **78.4** | **53.7** | **70.4** |

improvements (more than 20 pp) over the competitors. These large gains on minority classes indicate that successful pseudo label generation by OVIS and class-balanced training with CMB, enabling the segmentation DNN to learn discriminative features even for underrepresented object categories. In contrast, for dominant classes such as wall and floor, all methods achieve high IoU values (above 75%), and the margin of PLOVIS over the competitors is moderate.

**Table 3** shows the per-class IoU on S3DIS. PLOVIS achieves the highest IoU for 6 out of 13 classes. The improvement is most pronounced for sofa (+16 pp over ERDA), and table/bookcase (+4 pp over DG-Net). These furniture-related objects have visually distinct appearances that the OVIS model can recognize well even in the rendered point cloud images. On the other hand, PLOVIS does not achieve the highest IoU for certain classes such as column, window, door, and board, that are typically mounted on walls. We attribute this to the fact that the wall-mounted objects can be difficult for the OVIS model to distinguish from walls in the rendered images, making pseudo labels for these classes less reliable.

**Table 4** presents the per-class IoU for the two outdoor datasets. On Toronto3D, PLOVIS achieves the highest IoU for 5 out of 8 classes. The largest improvement is observed for fence, where PLOVIS achieves an IoU of 28.0%, while all competing methods remain below 3%. This drastic gap suggests that the competitors fail to learn a feature representation for fence from the sparse human-annotated labels in the training dataset. PLOVIS, in contrast, successfully supplements the scarce annotations with OVIS-based pseudo labels. On the other hand, for the class of road marking, PLOVIS performs significantly worse than full fine-tuning by a margin of 20 pp. This is probably because road markings occupy very small/thin regions on the ground surface and are difficult for the OVIS model to distinguish from the surrounding road in the rendered images. On Semantic3D, PLOVIS achieves the highest IoU for 4 out of 7 classes, demonstrating the positive impact of the proposed approach on these classes. Contrary to our expectations, the segmentation accuracy for the car class does not show significant improvement both for Toronto3D and Semantic3D. This is likely because vehicles captured by mobile laser-scanning systems or terrestrial laser scanners often have incomplete 3D shapes, which prevented the OVIS model from generating correct pseudo labels.

**Table 2.** Class-wise segmentation accuracy (IoU [%]) for the ScanNet dataset.

| Algorithms | wall | floor | cabinet | bed | chair | sofa | table | door | window | bookshelf | picture | counter | desk | curtain | refrigerator | shower curtain | toilet | sink | bathtub | other furniture |
|---|---|---|---|---|---|---|---|---|---|---|---|---|---|---|---|---|---|---|---|---|
| MLP probing | 80.1 | 92.1 | 37.8 | 62.9 | 77.0 | 57.1 | 59.0 | 56.5 | 58.6 | 16.5 | 14.9 | 16.6 | 40.0 | 67.7 | 41.1 | 64.4 | 84.5 | 26.4 | 73.6 | 30.4 |
| Decoder probing | 78.6 | 92.3 | 39.5 | 61.8 | 75.5 | 55.7 | 57.9 | 55.3 | 54.3 | 23.5 | 9.7 | 15.6 | 40.5 | 63.0 | 35.5 | 47.8 | 74.3 | 14.5 | 72.5 | 29.7 |
| Full fine-tuning | 75.9 | 92.6 | 37.1 | **66.3** | 73.9 | 52.2 | 56.8 | 46.7 | 50.9 | 22.0 | 7.8 | 19.5 | 36.4 | 57.5 | 31.3 | 39.5 | 68.7 | 17.3 | 62.8 | 24.9 |
| AAD-Net | 80.2 | 92.3 | 37.4 | 62.5 | 76.9 | 56.9 | 59.5 | 56.8 | 59.3 | 14.0 | 14.4 | 14.5 | 41.4 | 69.8 | 40.7 | 63.5 | **84.9** | 27.0 | **75.3** | 29.6 |
| DG-Net | 78.8 | 92.6 | 37.1 | 62.4 | 76.3 | 55.6 | 59.1 | 56.8 | 59.2 | 19.1 | 16.0 | 24.8 | 34.7 | 68.2 | 39.2 | 65.6 | 82.5 | 28.7 | 72.5 | 29.8 |
| ERDA | 79.5 | 92.7 | 38.8 | 62.9 | 77.7 | 58.0 | 57.4 | 56.9 | 60.2 | 20.1 | 15.9 | 20.7 | 34.2 | 71.8 | 41.2 | 64.7 | 82.5 | 22.6 | 68.5 | 30.0 |
| PLOVIS (ours) | **81.0** | **92.8** | **47.6** | 65.3 | **80.6** | **64.1** | **63.6** | **59.6** | **62.2** | **52.0** | **17.6** | **46.3** | **46.2** | **72.9** | **44.2** | **67.0** | 83.5 | **49.0** | 71.4 | **34.1** |

**Table 3.** Class-wise segmentation accuracy (IoU [%]) for the S3DIS dataset.

| Algorithms | ceiling | floor | wall | beam | column | window | door | table | chair | sofa | bookcase | board | clutter |
|---|---|---|---|---|---|---|---|---|---|---|---|---|---|
| MLP probing | 92.1 | 95.3 | 77.8 | 0.1 | **37.9** | 68.8 | 67.4 | 73.3 | 85.5 | 42.2 | 70.5 | 57.0 | 54.9 |
| Decoder probing | 91.5 | **96.7** | 79.0 | 0.1 | 33.0 | 64.6 | 62.7 | 71.9 | 83.8 | 42.6 | 68.8 | 55.8 | 50.0 |
| Full fine-tuning | 91.8 | **96.7** | **81.0** | **0.4** | 34.2 | 60.2 | 60.0 | 67.9 | 80.1 | 38.2 | 63.3 | **61.6** | 50.4 |
| AAD-Net | 91.8 | 95.3 | 78.6 | 0.1 | 36.1 | 68.2 | **67.9** | 73.0 | 85.3 | 47.8 | 70.3 | 56.5 | 53.0 |
| DG-Net | 91.5 | 95.4 | 77.8 | 0.1 | 34.8 | 67.6 | 69.6 | 73.3 | 84.8 | 50.7 | 71.2 | 55.1 | 52.6 |
| ERDA | 92.4 | 95.6 | 78.0 | 0.1 | 34.2 | **69.7** | 69.5 | 73.2 | 84.6 | 55.7 | 70.8 | 54.4 | 53.7 |
| PLOVIS (ours) | **93.2** | 95.7 | 79.0 | **0.4** | 37.0 | 67.5 | 66.3 | **77.1** | **87.4** | **72.0** | **75.0** | 52.5 | **55.5** |

**Table 4.** Class-wise segmentation accuracy (IoU [%]) for the Toronto3D and Semantic3D datasets.

| Algorithms | Toronto3D | | | | | | | | Semantic3D | | | | | | |
|---|---|---|---|---|---|---|---|---|---|---|---|---|---|---|---|
| | road | road marking | natural | building | utility line | pole | car | fence | man-made terrain | natural terrain | high vegetation | low vegetation | building | remaining hardscape | cars and trucks |
| MLP probing | 93.4 | 22.6 | 88.8 | 91.4 | 72.5 | 63.8 | 74.5 | 0.7 | 50.6 | 48.3 | 77.8 | **30.0** | 61.3 | 12.7 | **59.2** |
| Decoder probing | 93.7 | 29.7 | **89.9** | 91.5 | 68.2 | 63.7 | 68.2 | 1.7 | 51.6 | 59.7 | 71.0 | 22.0 | 57.5 | 12.7 | 44.7 |
| Full fine-tuning | 95.7 | **49.6** | 87.2 | 86.1 | 61.1 | 58.4 | 69.2 | 0.6 | 54.0 | 61.6 | 69.6 | 24.2 | 62.7 | **15.5** | 23.2 |
| AAD-Net | 94.0 | 18.9 | 88.5 | 91.2 | 72.6 | 63.4 | 74.7 | 1.7 | 48.7 | 54.0 | 75.5 | 26.7 | 59.3 | 13.0 | 59.5 |
| DG-Net | 94.2 | 20.9 | 88.4 | 89.9 | 72.0 | 62.5 | 72.5 | 0.9 | 49.7 | 54.7 | 74.9 | 27.8 | 61.2 | 13.6 | 56.8 |
| ERDA | 93.8 | 20.1 | 87.8 | 91.0 | 71.3 | 61.5 | **76.7** | 2.7 | 51.2 | 53.7 | 74.1 | 22.9 | 62.4 | 13.9 | 58.0 |
| PLOVIS (ours) | **94.3** | 29.2 | 89.1 | **91.6** | **73.0** | **65.9** | 75.4 | **28.0** | **57.9** | **67.2** | **81.0** | 29.0 | **70.6** | 15.3 | 54.7 |

In summary, PLOVIS consistently outperforms all competing algorithms when evaluated in mIoU and mACC, confirming the effectiveness of our proposed approach to data-efficient training. On the other hand, the class-wise evaluation reveals that pseudo-labeling remains challenging for object classes that are difficult for the OVIS model to recognize from rendered point cloud images.

Next, we evaluate the accuracy of each algorithm as a function of the number of point cloud scenes or the number of true labels per scene available for training. **Fig. 5** plots segmentation accuracy against the number of true labels per scene. In this experiment, the number of training scenes is fixed at the same values used in the aforementioned experiments, i.e., 60 for ScanNet, 40 for S3DIS, 20 for Toronto3D, and 15 for Semantic3D. Across all four datasets, PLOVIS consistently achieves the highest accuracy over the entire range of annotation budgets, indicating that the use of OVIS-based pseudo labels effectively compensates for the scarcity of manually annotated 3D points. In particular, on ScanNet and S3DIS, our method maintains relatively high accuracy even under the extremely sparse annotation (i.e., 20 labels per scene), whereas conventional methods, especially decoder probing and full fine-tuning, exhibit performance degradation due to overfitting.

**Figure 6** plots segmentation accuracy against the number of training 3D point clouds. In this experiment, the number of true labels per scene is fixed at 50 across all datasets. The proposed PLOVIS again shows the highest accuracy on all datasets, demonstrating that its advantage is maintained not only when pointwise annotations are scarce but also when the number of training scenes is very limited. The behavior of the competing methods is also consistent with **Fig. 5**. That is, a reduction in the number of training scenes renders these methods more vulnerable to overfitting.

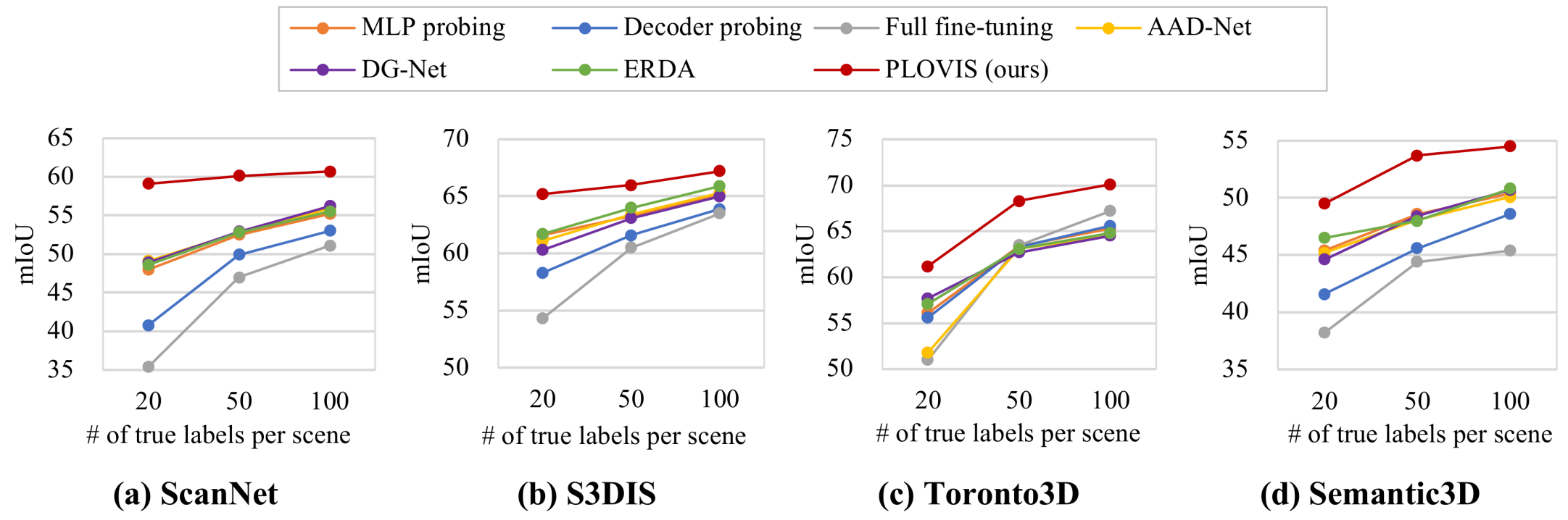

**Fig. 5.** Segmentation accuracy plotted against the number of human-annotated true labels per 3D point cloud scene. The number of training scenes is fixed at 60 for ScanNet, 40 for S3DIS, 20 for Toronto3D, and 15 for Semantic3D, respectively.

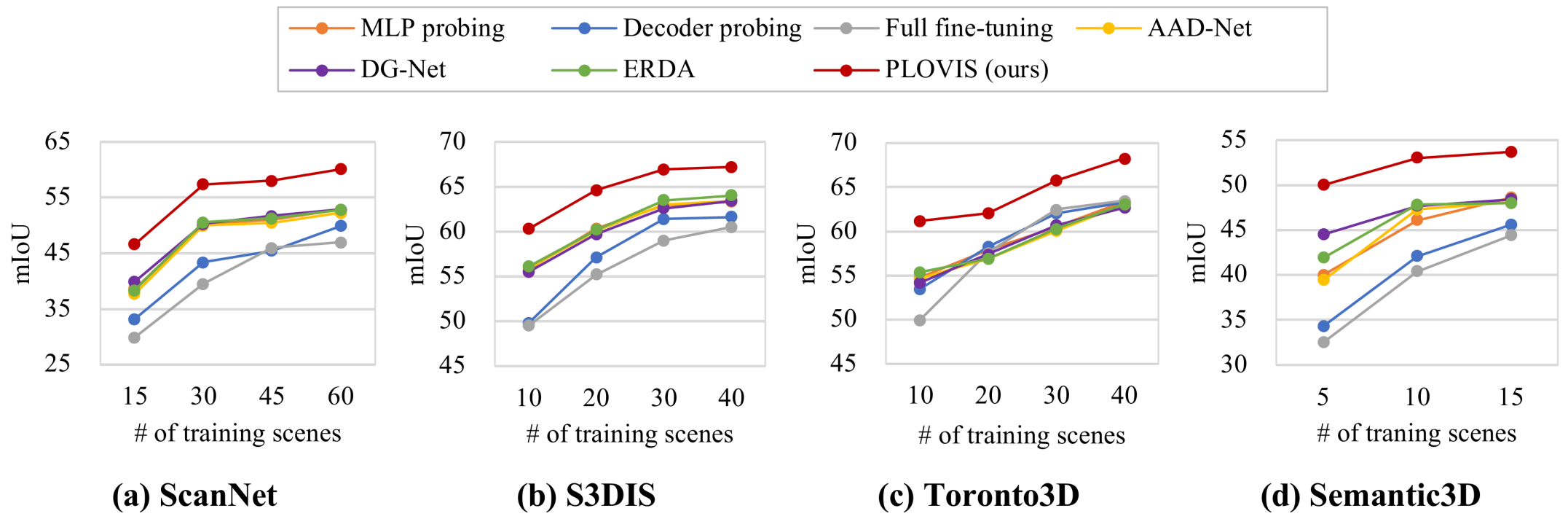

**Fig. 6.** Segmentation accuracy plotted against the number of training 3D point cloud scenes. The number of human-annotated true labels per scene is fixed at 50.

**Figure 7** shows qualitative comparison of segmentation results. We use 60 and 20 training scenes for ScanNet and Toronto3D respectively, each containing 50 true labels. ERDA is used as a competitor since it employs the same approach as PLOVIS, i.e., pseudo labeling, for training. The major differences between the two methods are highlighted by pink dashed boxes in **Fig. 7**. Specifically, in the indoor scene, ERDA incorrectly classifies a bookshelf as a cabinet, whereas PLOVIS correctly identifies it as a bookshelf. Similarly, in the outdoor scene, PLOVIS successfully segments the fence in the roadside. The improvements of these categories are attributed to OVIS-based pseudo-labeling, which is consistent with the class-wise results reported in **Tables 2** and **Table 4**.

#### 4.2.2. In-depth evaluation of PLOVIS

This subsection evaluates the validity of the design choices underlying the proposed algorithm. In the experiments presented in this subsection, we use 60 training scenes for ScanNet, 40 for S3DIS, 20 for Toronto3D, and 15 for Semantic3D, with the number of true labels per scene set to 50.

**Two-stage pseudo label filtering. Table 5** demonstrates the effectiveness of the proposed two-stage filtering of pseudo labels. In the table, case 1 applies neither the first-stage nor the second-stage filtering, using all pseudo labels generated by the OVIS model for training. Case 1 already outperforms MLP probing on three out of four datasets, confirming that incorporating OVIS-based pseudo labels is beneficial even without any filtering. However, the gains remain limited because noisy pseudo labels contaminate the supervisory signal. Case 2 enables only the first stage (filtering by confidence), and case 3 enables only the second stage (filtering by loss). Each filtering stage independently contributes to accuracy improvement.

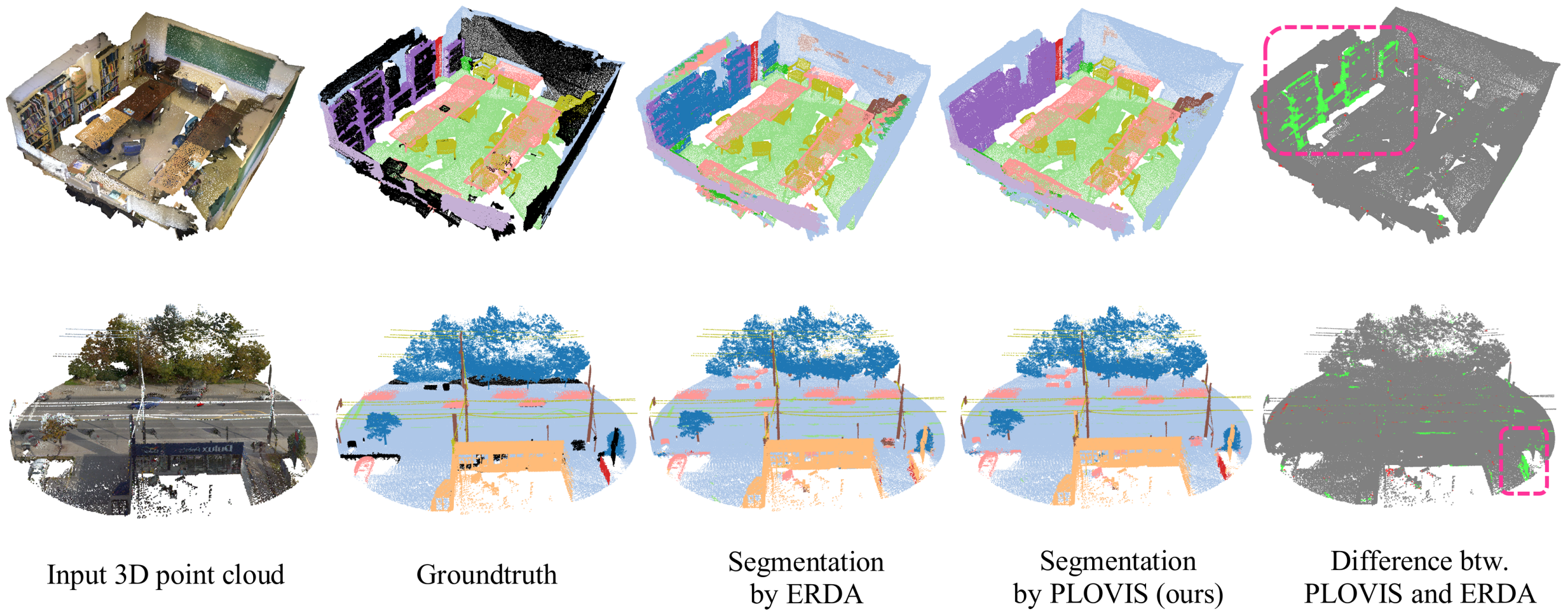

**Fig. 7.** Segmentation results for indoor (ScanNet) and outdoor (Toronto3D) scenes. In the groundtruth and segmentation results, point colors indicate semantic categories, while black represents unlabeled points excluded from evaluation. The difference images show improvements by PLOVIS. That is, green points indicate correct predictions by PLOVIS but incorrect predictions by ERDA, whereas red points indicate correct predictions by ERDA but incorrect predictions by PLOVIS. Major improvements are highlighted by pink dashed boxes.

**Table 5.** Effectiveness of two-stage filtering of pseudo labels (mIoU [%]).

| | First stage: Filtering by confidence | Second stage: Filtering by loss | ScanNet | S3DIS | Toronto3D | Semantic3D |
|---|---|---|---|---|---|---|
| Case 1 | No | No | 57.5 | 61.9 | 66.1 | 49.8 |
| Case 2 | Yes | No | 58.2 | 65.5 | 67.1 | 52.9 |
| Case 3 | No | Yes ($G^j_{small}$) | 59.6 | 64.0 | 67.4 | 51.5 |
| Case 4 (ours) | Yes | Yes ($G^j_{small}$) | **60.1** | **66.0** | **68.3** | **53.7** |
| Case 5 | Yes | Yes ($G^j_{large}$) | 55.8 | 64.4 | 59.5 | 51.9 |

These results suggest that the filtering succeeds, to a certain extent, in removing low-confidence pseudo labels and presumably incorrect pseudo labels. Case 4, which is our proposed configuration, applies both stages and achieves the highest segmentation accuracy on all four datasets. The consistent improvements of case 4 over cases 2 and 3 demonstrate that sequentially applying the two filtering stages yields additive benefits, validating our two-stage design. Furthermore, case 5 examines the design of the second stage filtering. In case 5, pseudo labels whose losses belong to $G^j_{large}$, rather than $G^j_{small}$, are appended to the CMB. This configuration retains pseudo labels with higher loss values, which are more likely to be incorrectly pseudo-labeled. As a result, case 5 leads to significant accuracy degradation on all datasets compared to case 4. This result confirms that the loss values computed by the DNN during training serve as a reasonably effective measure of pseudo label error, and that retaining pseudo labels with small loss values is essential for effective training.

We quantitatively verify that the two-stage filtering improves the quality of pseudo labels. To this end, we define two evaluation metrics: label inconsistency (LI) and label accuracy (LA). LI measures the disagreement between the pseudo label vector $\tilde{\mathbf{y}}_i$ generated by the OVIS model and the ground-truth label vector, computed as the cross-entropy between the two vectors. A smaller LI value indicates that the OVIS model prediction is closer to the true label. LA, on the other hand, measures the agreement rate between the class corresponding to the maximum element in the pseudo label vector $\tilde{\mathbf{y}}_i$ and the ground-truth class. A larger LA value indicates more accurate pseudo labels. We can compute LI and LA since the point clouds used in our experiments are originally densely annotated. **Table 6** reports the mean LI and LA values computed around 100th epoch during PLOVIS training. In the table, $\mathcal{D}_0 \setminus \mathcal{D}_1$ denotes the set of pseudo labels removed at the first stage (red points in **Fig. 4**), $\mathcal{D}_1 \setminus \mathcal{D}_2$ denotes the set of pseudo labels removed at the second stage (blue points in **Fig. 4**), and $\mathcal{D}_2$ denotes the set of pseudo labels appended to the CMB and used for training (green points in **Fig. 4**). As each filtering stage is applied, LI consistently decreases and LA consistently increases across all datasets, indicating that the pseudo labels are progressively "purified" through the two-stage filtering. Specifically, evaluation by LA shows that nearly 90% of pseudo labels after the two-stage filtering are correct on ScanNet, S3DIS, and Toronto3D. On Semantic3D, the absolute LA value is lower compared to the other datasets. This is probably because the OVIS model finds it more challenging to generate accurate pseudo labels for the outdoor scenes in Semantic3D, which contain objects with more complex geometry and appearance. Nevertheless, the filtering still produces a meaningful improvement in label quality even on this challenging dataset.

**Joint training with true labels and pseudo labels.** PLOVIS trains the segmentation head using sparse true labels and dense pseudo labels. In the preceding experiments, we fixed the coefficient $w$, which balances the true label loss $L_{true}$ and the pseudo label loss $L_{pseudo}$, at 0.5. Here, we vary $w$ to investigate its effect on segmentation accuracy. **Fig. 8** shows the relationship between $w$ and mIoU on each dataset. Setting $w = 1$ uses true labels only, while $w = 0$ uses pseudo labels only for training. Across all four datasets, peak accuracy appears at an intermediate value of $w$ (e.g., 0.5). This result confirms the benefit of combining human-annotated true labels and OVIS-generated pseudo labels under data-scarce conditions.

**Zero-shot learning.** Setting $w = 0$ in our loss function can be interpreted as zero-shot learning, where no human annotations are used for training. We compare PLOVIS under the zero-shot setting with existing zero-shot approaches to open-vocabulary 3D point cloud segmentation. OpenScene [56], GeoZe [63], and OV3D [54] all achieve open-vocabulary point cloud segmentation by embedding 3D point clouds and their source 2D image sequences into the CLIP latent space. As shown in **Table 7**, PLOVIS under the zero-shot setting achieves an mIoU of 51.6% and an mAcc of 68.2% on ScanNet, surpassing several, though not all, of the existing methods. It is worth noting that, unlike the existing algorithms ([54], [56], [63]), PLOVIS does not require the 2D image sequences from which the 3D point clouds were reconstructed. The compared algorithms presuppose the availability of image sequences to jointly embed 3D point clouds and their corresponding natural images into the CLIP latent space. PLOVIS achieves accuracy comparable to some existing methods with substantially higher data efficiency. The favorable zero-shot learning capability of PLOVIS are attributed to the adoption of the powerful PCF model (Sonata) and OVIS model (DeCLIP) as well as the proposed pseudo label filtering mechanism.

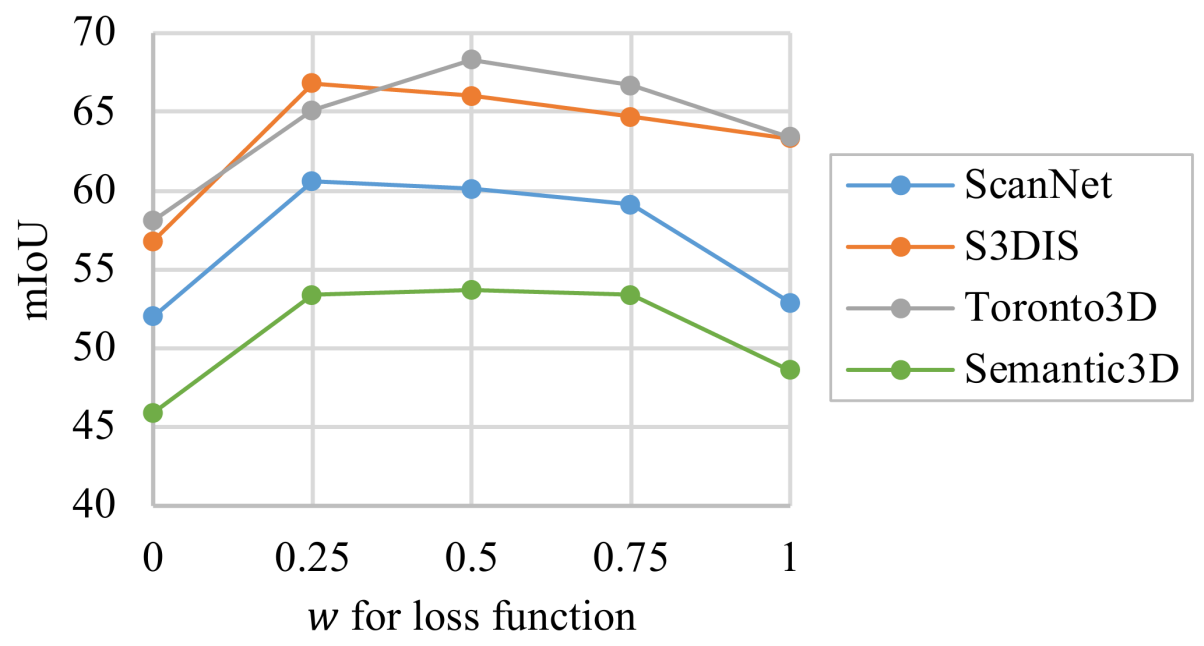

**Fig. 8.** Segmentation accuracy plotted against the balancing coefficient $w$ used in our loss function.

**Table 6.** Quality of pseudo labels measured in label inconsistency (LI) and label accuracy (LA [%]).

| Pseudo labels | ScanNet | | S3DIS | | Toronto3D | | Semantic3D | |
|---|---|---|---|---|---|---|---|---|
| | LI ↓ | LA ↑ | LI ↓ | LA ↑ | LI ↓ | LA ↑ | LI ↓ | LA ↑ |
| $\mathcal{D}_0 \setminus \mathcal{D}_1$ (removed at the first stage) | 5.13 | 48.3 | 4.10 | 58.5 | 3.57 | 53.3 | 5.33 | 61.0 |
| $\mathcal{D}_1 \setminus \mathcal{D}_2$ (removed at the second stage) | 2.94 | 82.4 | 1.61 | 91.1 | 1.41 | 90.4 | 4.66 | 68.1 |
| $\mathcal{D}_2$ (used for training) | 2.24 | 86.7 | 1.34 | 92.7 | 1.29 | 91.2 | 4.08 | 72.7 |

**Table 7.** Comparison with existing algorithms for open-vocabulary semantic segmentation of 3D point clouds.

| Algorithms | Training requires 2D image sequence? | ScanNet | |
|---|---|---|---|
| | | mIoU [%] | mAcc [%] |
| OpenScene + OpenSeg | Yes | 47.5 | 70.7 |
| OpenScene + LSeg | Yes | 54.2 | 66.6 |
| GeoZe + OpenSeg | Yes | 45.6 | — |
| GeoZe + LSeg | Yes | 55.8 | — |
| OV3D | Yes | 57.3 | 72.9 |
| PLOVIS with $w = 0$ | No | 51.6 | 68.2 |

## 5. Conclusion and future work

This paper tackles semantic segmentation of 3D point clouds under data insufficiency conditions that frequently arise in real-world applications: scarcity of training scenes, scarcity of annotations, and absence of image sequences. To overcome these challenges, we propose *Point pseudo-Labeling via Open-Vocabulary Image Segmentation* (*PLOVIS*), a data-efficient training algorithm that leverages an Open-Vocabulary Image Segmentation (OVIS) model as a pseudo label generator. PLOVIS renders training 3D point clouds into 2D images and obtains pointwise pseudo labels via zero-shot pixel-level prediction by the OVIS model. To cope with the noise and class imbalance inherent in the generated pseudo labels, PLOVIS employs a two-stage filtering mechanism based on prediction confidence and training loss, along with a class-balanced memory bank. A lightweight segmentation head is jointly trained using both sparse human-annotated true labels and dense pseudo labels, while the point cloud foundation model and the OVIS model remain frozen to mitigate overfitting. Experiments on four benchmark datasets under data-scarce conditions demonstrated that PLOVIS consistently outperforms standard fine-tuning strategies and state-of-the-art weakly supervised learning methods across all datasets. This result confirms that the crux of our approach, i.e., pseudo-labeling by an OVIS model, is effective for data-efficient training of a 3D point cloud segmentation model.

There are several directions for future work. For example:

- Improving pseudo label quality. Future work should explore two promising directions. First, the partially corrupted and pixelated appearance of rendered point cloud images degrades the recognition accuracy of the OVIS model. Restoring such renderings into more photorealistic images using image generation models (e.g., diffusion-based inpainting) prior to OVIS inference could improve pseudo label quality. Second, the current implementation uses simple text prompts constructed by concatenating a fixed template (i.e., "a photo of a") with a class name, which may not always yield optimal text embeddings for the OVIS model. Systematic prompt engineering such as incorporating contextual descriptions, attribute-rich class definitions, or learned prompt optimization, could further boost the accuracy of zero-shot pseudo-labeling by the OVIS model.
- Expanding applicability to diverse tasks and domains. The current PLOVIS framework focuses on closed-set semantic segmentation of indoor and outdoor 3D point clouds. Extending its applicability to other 3D scene understanding tasks, such as instance and panoptic segmentation, as well as to point clouds acquired by different sensor modalities (e.g., LiDAR and RGB-D), and to specialized domains such as industrial facilities and construction sites, constitutes a promising avenue for future research.

#### Acknowledgements

This work was supported by the Japan Society for the Promotion of Science (JSPS) KAKENHI (Grant No. 24K14992).

## References:

[1] Y. Sun, X. Zhang, and Y. Miao, "A review of point cloud segmentation for understanding 3D indoor scenes," Visual Intelligence, Vol. 2, Article No. 14, 2024. https://doi.org/10.1007/s44267-024-00046-x

[2] S. Sarker, P. Sarker, G. Stone, R. Gorman, A. Tavakkoli, G. Bebis, and J. Sattarvand, "A comprehensive overview of deep learning techniques for 3D point cloud classification and semantic segmentation," Machine Vision and Applications, Vol. 35, Article No. 67, 2024. https://doi.org/10.1007/s00138-024-01543-1

[3] X. Wu, L. Jiang, P.-S. Wang, Z. Liu, X. Liu, Y. Qiao, W. Ouyang, T. He, and H. Zhao, "Point Transformer V3: Simpler, faster, stronger," Proceedings of the IEEE/CVF Conference on Computer Vision and Pattern Recognition, pp. 4840-4851, 2024. https://doi.org/10.1109/CVPR52733.2024.00463

[4] X. Wu, D. DeTone, D. Frost, T. Shen, C. Xie, N. Yang, J. Engel, R. Newcombe, H. Zhao, and J. Straub, "Sonata: Self-supervised learning of reliable point representations," Proceedings of the IEEE/CVF Conference on Computer Vision and Pattern Recognition, pp. 22193-22204, 2025. https://doi.org/10.1109/CVPR52734.2025.02067

[5] A. Nekrasov, J. Schult, O. Litany, B. Leibe, and F. Engelmann, "Mix3D: Out-of-context data augmentation for 3D scenes," Proceedings of the International Conference on 3D Vision, pp. 116-125, 2021. https://doi.org/10.1109/3DV53792.2021.00022

[6] A. Dai, A. X. Chang, M. Savva, M. Halber, T. Funkhouser, and M. Nießner, "ScanNet: Richly-annotated 3D reconstructions of indoor scenes," Proceedings of the IEEE Conference on Computer Vision and Pattern Recognition, pp. 2432-2443, 2017. https://doi.org/10.1109/CVPR.2017.261

[7] I. Armeni, O. Sener, A. R. Zamir, H. Jiang, I. Brilakis, M. Fischer, and S. Savarese, "3D semantic parsing of large-scale indoor spaces," Proceedings of the IEEE Conference on Computer Vision and Pattern Recognition, pp. 1534-1543, 2016. https://doi.org/10.1109/CVPR.2016.170

[8] H. Caesar, V. Bankiti, A. H. Lang, S. Vora, V. E. Liong, Q. Xu, A. Krishnan, Y. Pan, G. Baldan, and O. Beijbom, "nuScenes: A multimodal dataset for autonomous driving," Proceedings of the IEEE/CVF Conference on Computer Vision and Pattern Recognition, pp. 11618-11628, 2020. https://doi.org/10.1109/CVPR42600.2020.01164

[9] X. Yan, J. Gao, C. Zheng, C. Zheng, R. Zhang, S. Cui, and Z. Li, "2DPASS: 2D priors assisted semantic segmentation on LiDAR point clouds," Proceedings of the European Conference on Computer Vision, pp. 677-695, 2022. https://doi.org/10.1007/978-3-031-19815-1_39

[10] S. Dong, F. Liu, R. Yao, and G. Lin, "Leveraging large-scale pretrained vision foundation models for label-efficient 3D point cloud segmentation," Proceedings of the International Conference on Image and Graphics, pp. 304-315, 2026. https://doi.org/10.1007/978-981-95-3393-0_25

[11] K. Knaebel, K. Yilmaz, D. de Geus, A. Hermans, D. Adrian, T. Linder, and B. Leibe, "DINO in the room: Leveraging 2D foundation models for 3D segmentation," Proceedings of the International Conference on 3D Vision, pp. 1884-1894, 2026. https://doi.org/10.1109/3DV69130.2026.00178

[12] A. Xiao, X. Zhang, L. Shao, and S. Lu, "A survey of label-efficient deep learning for 3D point clouds," IEEE Transactions on Pattern Analysis and Machine Intelligence, Vol. 46, No. 12, pp. 9139-9160, 2024. https://doi.org/10.1109/TPAMI.2024.3416302

[13] J. Wang, Y. Liu, H. Tan, and M. Zhang, "A survey on weakly supervised 3D point cloud semantic segmentation," IET Computer Vision, Vol. 18, No. 3, pp. 329-342, 2024. https://doi.org/10.1049/cvi2.12250

[14] F. Langer, A. Milioto, A. Haag, J. Behley, and C. Stachniss, "Domain transfer for semantic segmentation of LiDAR data using deep neural networks," Proceedings of the IEEE/RSJ International Conference on Intelligent Robots and Systems, pp. 8263-8270, 2020. https://doi.org/10.1109/IROS45743.2020.9341508

[15] L. Yi, B. Gong, and T. Funkhouser, "Complete & label: A domain adaptation approach to semantic segmentation of LiDAR point clouds," Proceedings of the IEEE/CVF Conference on Computer Vision and Pattern Recognition, pp. 15358-15368, 2021. https://doi.org/10.1109/CVPR46437.2021.01511

[16] J. Wang, B. Chen, Y. Li, B. Kang, Y. Chen, and Z. Tian, "DeCLIP: Decoupled learning for open-vocabulary dense perception," Proceedings of the IEEE/CVF Conference on Computer Vision and Pattern Recognition, pp. 14824-14834, 2025. https://doi.org/10.1109/CVPR52734.2025.01381

[17] W. Tan, N. Qin, L. Ma, Y. Li, J. Du, G. Cai, K. Yang, and J. Li, "Toronto-3D: A large-scale mobile LiDAR dataset for semantic segmentation of urban roadways," Proceedings of the IEEE/CVF Conference on Computer Vision and Pattern Recognition Workshops, pp. 202-203, 2020. https://doi.org/10.1109/CVPRW50498.2020.00109

[18] X. Ma, C. Qin, H. You, H. Ran, and Y. Fu, "Rethinking network design and local geometry in point cloud: A simple residual MLP framework," Proceedings of the International Conference on Learning Representations, 2022. https://doi.org/10.48550/arXiv.2202.07123

[19] G. Qian, Y. Li, H. Peng, J. Mai, H. Hammoud, M. Elhoseiny, and B. Ghanem, "PointNeXt: Revisiting PointNet++ with improved training and scaling strategies," Advances in Neural Information Processing Systems, Vol. 35, Article No. 1685, pp. 23192-23204, 2022. https://doi.org/10.52202/068431-1685

[20] P.-S. Wang, "OctFormer: Octree-based transformers for 3D point clouds," ACM Transactions on Graphics, Vol. 42, No. 4, Article No. 155, 2023. https://doi.org/10.1145/3592131

[21] J. Behley, M. Garbade, A. Milioto, J. Quenzel, S. Behnke, C. Stachniss, and J. Gall, "SemanticKITTI: A dataset for semantic scene understanding of LiDAR sequences," Proceedings of the IEEE/CVF International Conference on Computer Vision, pp. 9297-9307, 2019. https://doi.org/10.1109/ICCV.2019.00939

[22] L. Tang, Z. Chen, S. Zhao, C. Wang, and D. Tao, "All points matter: Entropy-regularized distribution alignment for weakly supervised 3D segmentation," Advances in Neural Information Processing Systems, Vol. 36, Article No. 3440, pp. 78657-78673, 2023. https://doi.org/10.52202/075280-3440

[23] Y. Wu, Z. Yan, S. Cai, G. Li, X. Han, and S. Cui, "PointMatch: A consistency training framework for weakly supervised semantic segmentation of 3D point clouds," Computers & Graphics, Vol. 116, pp. 427-436, 2023. https://doi.org/10.1016/j.cag.2023.09.006

[24] Z. Wu, Y. Wu, G. Lin, and J. Cai, "Reliability-adaptive consistency regularization for weakly supervised point cloud segmentation," International Journal of Computer Vision, Vol. 132, No. 6, pp. 2276-2289, 2024. https://doi.org/10.1007/s11263-023-01975-8

[25] Z. Pan, W. Gao, S. Liu, and G. Li, "Distribution guidance network for weakly supervised point cloud semantic segmentation," Advances in Neural Information Processing Systems, Vol. 37, pp. 32400-32420, 2024. https://doi.org/10.52202/079017-1018

[26] Z. Pan, N. Zhang, W. Gao, S. Liu, and G. Li, "Point cloud semantic segmentation with sparse and inhomogeneous annotations," Proceedings of the AAAI Conference on Artificial Intelligence, Vol. 39, No. 6, Article No. 707, pp. 6354-6362, 2025. https://doi.org/10.1609/aaai.v39i6.32680

[27] K. Sohn, D. Berthelot, C.-L. Li, Z. Zhang, N. Carlini, E. D. Cubuk, A. Kurakin, H. Zhang, and C. Raffel, "FixMatch: Simplifying semi-supervised learning with consistency and confidence," Advances in Neural Information Processing Systems, Vol. 33, Article No. 51, pp. 596-608, 2020. https://doi.org/10.48550/arXiv.2001.07685

[28] G. Li, G. Kang, X. Wang, Y. Wei, and Y. Yang, "Adversarially masking synthetic to mimic real: Adaptive noise injection for point cloud segmentation adaptation," Proceedings of the IEEE/CVF Conference on Computer Vision and Pattern Recognition, pp. 20464-20474, 2023. https://doi.org/10.1109/CVPR52729.2023.01960

[29] A. Xiao, J. Huang, K. Liu, D. Guan, X. Zhang, and S. Lu, "Domain adaptive LiDAR point cloud segmentation via density-aware self-training," IEEE Transactions on Intelligent Transportation Systems, Vol. 25, No. 10, pp. 13627-13639, 2024. https://doi.org/10.1109/TITS.2024.3386865

[30] M. Oquab, T. Darcet, T. Moutakanni, H. Vo, M. Szafraniec, V. Khalidov, P. Fernandez, D. Haziza, F. Massa, A. El-Nouby, M. Assran, N. Ballas, W. Galuba, R. Howes, P.-Y. Huang, S.-W. Li, I. Misra, M. Rabbat, V. Sharma, G. Synnaeve, H. Xu, H. Jégou, J. Mairal, P. Labatut, A. Joulin, and P. Bojanowski, "DINOv2: Learning robust visual features without supervision," Transactions on Machine Learning Research, 2024. https://doi.org/10.48550/arXiv.2304.07193

[31] H. Kweon, J. Kim, and K.-J. Yoon, "Weakly supervised point cloud semantic segmentation via artificial oracle," Proceedings of the IEEE/CVF Conference on Computer Vision and Pattern Recognition, pp. 3721-3731, 2024. https://doi.org/10.1109/CVPR52733.2024.00357

[32] J. Deng, J. Lu, and T. Zhang, "Quantity-quality enhanced self-training network for weakly supervised point cloud semantic segmentation," IEEE Transactions on Pattern Analysis and Machine Intelligence, Vol. 47, No. 5, pp. 3580-3596, 2025. https://doi.org/10.1109/TPAMI.2025.3532637

[33] A. Kirillov, E. Mintun, N. Ravi, H. Mao, C. Rolland, L. Gustafson, T. Xiao, S. Whitehead, A. C. Berg, W.-Y. Lo, P. Dollár, and R. Girshick, "Segment Anything," Proceedings of the IEEE/CVF International Conference on Computer Vision, pp. 4015-4026, 2023. https://doi.org/10.1109/ICCV51070.2023.00371

[34] A. Radford, J. W. Kim, C. Hallacy, A. Ramesh, G. Goh, S. Agarwal, G. Sastry, A. Askell, P. Mishkin, J. Clark, G. Krueger, and I. Sutskever, "Learning transferable visual models from natural language supervision," Proceedings of the International Conference on Machine Learning, 2021. https://doi.org/10.48550/arXiv.2103.00020

[35] H. Kweon and K.-J. Yoon, "Joint learning of 2D-3D weakly supervised semantic segmentation," Advances in Neural Information Processing Systems, Vol. 35, Article No. 2211, pp. 30499-30511, 2022. https://doi.org/10.52202/068431-2211

[36] L. Duan, S. Zhao, X. Weng, J. Zhang, and G.-S. Xia, "High-quality pseudo-labeling for point cloud segmentation with scene-level annotation," IEEE Transactions on Pattern Analysis and Machine Intelligence, Vol. 47, No. 10, pp. 9360-9366, 2025. https://doi.org/10.1109/TPAMI.2025.3583071

[37] T. Gao, X. Yao, and D. Chen, "SimCSE: Simple contrastive learning of sentence embeddings," Proceedings of the Conference on Empirical Methods in Natural Language Processing, pp. 6894-6910, 2021. https://doi.org/10.18653/v1/2021.emnlp-main.552

[38] A. Xiao, J. Huang, D. Guan, X. Zhang, S. Lu, and L. Shao, "Unsupervised point cloud representation learning with deep neural networks: A survey," IEEE Transactions on Pattern Analysis and Machine Intelligence, Vol. 45, No. 9, pp. 11321-11339, 2023. https://doi.org/10.1109/TPAMI.2023.3262786

[39] X. Yu, L. Tang, Y. Rao, T. Huang, J. Zhou, and J. Lu, "Point-BERT: Pre-training 3D point cloud transformers with masked point modeling," Proceedings of the IEEE/CVF Conference on Computer Vision and Pattern Recognition, pp. 19313-19322, 2022. https://doi.org/10.1109/CVPR52688.2022.01871

[40] Y. Pang, W. Wang, F. E. H. Tay, W. Liu, Y. Tian, and L. Yuan, "Masked autoencoders for point cloud self-supervised learning," Proceedings of the European Conference on Computer Vision, pp. 604-621, 2022. https://doi.org/10.1007/978-3-031-20086-1_35

[41] T. Furuya, "MaskLRF: Self-supervised pretraining via masked autoencoding of local reference frames for rotation-invariant 3D point set analysis," IEEE Access, Vol. 12, pp. 73340-73353, 2024. https://doi.org/10.1109/ACCESS.2024.3404016

[42] X. Wu, X. Wen, X. Liu, and H. Zhao, "Masked scene contrast: A scalable framework for unsupervised 3D representation learning," Proceedings of the IEEE/CVF Conference on Computer Vision and Pattern Recognition, pp. 9415-9424, 2023. https://doi.org/10.1109/CVPR52729.2023.00908

[43] Y. Zhang, X. Wu, Y. Lao, C. Wang, Z. Tian, N. Wang, and H. Zhao, "Concerto: Joint 2D-3D self-supervised learning emerges spatial representations," Advances in Neural Information Processing Systems, Vol. 38, pp. 77509-77533, 2025. https://doi.org/10.52202/085713-2338

[44] M. Xing, X. Wang, and Y. Shi, "DoReMi: A domain-representation mixture framework for generalizable 3D understanding," unpublished, arXiv:2511.11232, 2025. https://doi.org/10.48550/arXiv.2511.11232

[45] B. Yang, M. Abdelsamad, M. Zhang, and A. P. Condurache, "Towards foundation models for 3D scene understanding: Instance-aware self-supervised learning for point clouds," Proceedings of the IEEE/CVF Conference on Computer Vision and Pattern Recognition, pp. 2938-2947, 2026. https://doi.org/10.48550/arXiv.2603.25165

[46] J. Wu, X. Li, S. Xu, H. Yuan, H. Ding, Y. Yang, X. Li, J. Zhang, Y. Tong, X. Jiang, B. Ghanem, and D. Tao, "Towards open vocabulary learning: A survey," IEEE Transactions on Pattern Analysis and Machine Intelligence, Vol. 46, No. 7, pp. 5092-5113, 2024. https://doi.org/10.1109/TPAMI.2024.3361862

[47] C. Zhu and L. Chen, "A survey on open-vocabulary detection and segmentation: Past, present, and future," IEEE Transactions on Pattern Analysis and Machine Intelligence, Vol. 46, No. 12, pp. 8954-8975, 2024. https://doi.org/10.1109/TPAMI.2024.3413013

[48] B. Li, K. Q. Weinberger, S. Belongie, V. Koltun, and R. Ranftl, "Language-driven semantic segmentation," Proceedings of the International Conference on Learning Representations, 2022. https://doi.org/10.48550/arXiv.2201.03546

[49] G. Ghiasi, X. Gu, Y. Cui, and T.-Y. Lin, "Scaling open-vocabulary image segmentation with image-level labels," Proceedings of the European Conference on Computer Vision, pp. 540-557, 2022. https://doi.org/10.1007/978-3-031-20059-5_31

[50] F. Liang, B. Wu, X. Dai, K. Li, Y. Zhao, H. Zhang, P. Zhang, P. Vajda, and D. Marculescu, "Open-vocabulary semantic segmentation with Mask-adapted CLIP," Proceedings of the IEEE/CVF Conference on Computer Vision and Pattern Recognition, pp. 7061-7070, 2023. https://doi.org/10.1109/CVPR52729.2023.00682

[51] S. Cho, H. Shin, S. Hong, A. Arnab, P. H. Seo, and S. Kim, "CAT-Seg: Cost aggregation for open-vocabulary semantic segmentation," Proceedings of the IEEE/CVF Conference on Computer Vision and Pattern Recognition, pp. 4113-4123, 2024. https://doi.org/10.1109/CVPR52733.2024.00394

[52] S. Xiao, R. Kabra, Y. Li, D. Lee, J. Carreira, and P. Panda, "OpenWorldSAM: Extending SAM2 for universal image segmentation with language prompts," Advances in Neural Information Processing Systems, Vol. 38, pp. 39943-39968, 2025. https://doi.org/10.52202/085713-1193

[53] N. Carion et al., "SAM 3: Segment Anything with Concepts," Proceedings of the International Conference on Learning Representations, 2026. https://doi.org/10.48550/arXiv.2511.16719

[54] L. Jiang, S. Shi, and B. Schiele, "Open-vocabulary 3D semantic segmentation with foundation models," Proceedings of the IEEE/CVF Conference on Computer Vision and Pattern Recognition, pp. 21284-21294, 2024. https://doi.org/10.1109/CVPR52733.2024.02011

[55] Q. He, J. Peng, Z. Jiang, K. Wu, X. Ji, J. Zhang, Y. Wang, C. Wang, M. Chen, and Y. Wu, "UniM-OV3D: Uni-modality open-vocabulary 3D scene understanding with fine-grained feature representation," Proceedings of the International Joint Conference on Artificial Intelligence, Article No. 90, pp. 812-820, 2024. https://doi.org/10.24963/ijcai.2024/90

[56] S. Peng, K. Genova, C. Jiang, A. Tagliasacchi, M. Pollefeys, and T. Funkhouser, "OpenScene: 3D scene understanding with open vocabularies," Proceedings of the IEEE/CVF Conference on Computer Vision and Pattern Recognition, pp. 815-824, 2023. https://doi.org/10.1109/CVPR52729.2023.00085

[57] E. Arazo, D. Ortego, P. Albert, N. E. O'Connor, and K. McGuinness, "Unsupervised label noise modeling and loss correction," Proceedings of the International Conference on Machine Learning, pp. 312-321, 2019. https://doi.org/10.48550/arXiv.1904.11238

[58] J. Li, R. Socher, and S. C. H. Hoi, "DivideMix: Learning with noisy labels as semi-supervised learning," Proceedings of the International Conference on Learning Representations, 2020. https://doi.org/10.48550/arXiv.2002.07394

[59] F. R. Cordeiro, V. Belagiannis, I. Reid, and G. Carneiro, "PropMix: Hard sample filtering and proportional MixUp for learning with noisy labels," Proceedings of the British Machine Vision Conference, Article No. 264, 2021. https://doi.org/10.5244/C.35.264

[60] S. Ioffe and C. Szegedy, "Batch normalization: Accelerating deep network training by reducing internal covariate shift," Proceedings of the International Conference on Machine Learning, pp. 448-456, 2015. https://doi.org/10.48550/arXiv.1502.03167

[61] I. Loshchilov and F. Hutter, "Decoupled weight decay regularization," Proceedings of the International Conference on Learning Representations, 2019. https://doi.org/10.48550/arXiv.1711.05101

[62] T. Hackel, N. Savinov, L. Ladický, J. D. Wegner, K. Schindler, and M. Pollefeys, "Semantic3D.net: A new large-scale point cloud classification benchmark," ISPRS Annals of the Photogrammetry, Remote Sensing and Spatial Information Sciences, Vol. IV-1/W1, pp. 91-98, 2017. https://doi.org/10.5194/isprs-annals-IV-1-W1-91-2017

[63] G. Mei, L. Riz, Y. Wang, and F. Poiesi, "Geometrically-driven aggregation for zero-shot 3D point cloud understanding," Proceedings

of the IEEE/CVF Conference on Computer Vision and Pattern Recognition, pp. 27896-27905, 2024. https://doi.org/10.1109/CVPR52733.2024.02635